%% file: main.tex
\documentclass[graybox]{svmult}

\input{packages}


\newcommand{\buildbook}{false}
\newcommand{\isdraft}{true}

\makeindex             

\makeglossary

\begin{document}

\ifthenelse{\equal{true}{\buildbook}}{
\title{Contribution Title}
}
{
\title*{Intra and Cross-spectrum Iris Presentation Attack Detection in the NIR and Visible Domains}
}

\titlerunning{Intra and Cross-spectrum Iris Presentation Attack Detection}

\author{Meiling Fang, Fadi Boutros, Naser Damer}


\institute{Meiling Fang \at Fraunhofer Institute for Computer Graphics Research IGD,
Darmstadt, Germany \email{meiling.fang@igd.fraunhofer.de}
\and Fadi Boutros \at Fraunhofer Institute for Computer Graphics Research IGD,
Darmstadt, Germany  \email{fadi.boutros@igd.fraunhofer.de}
\and Naser Damer \at Fraunhofer Institute for Computer Graphics Research IGD,
Darmstadt, Germany  \email{naser.damer@igd.fraunhofer.de}}

%

%
\maketitle

\ifthenelse{\equal{false}{\isdraft}}{
\texttt{(filename: chapterXX/main.tex)}\\
\texttt{(corresponding author: \textbf{Meiling Fang})}\\
\texttt{(authors: \textbf{Meiling Fang, Fadi Boutros, Naser Damer})}\\
\texttt{(status: \textcolor{red}{\bf EMPTY} [NO])}\\
\texttt{(status: \textcolor{blue}{\bf DRAFT} [YES])}\\
\texttt{(status: \textcolor{green}{\bf COMPLETED} [NO])}\\
\texttt{(action required: \textbf{to write})}\\
}


\abstract*{
Iris Presentation Attack Detection (PAD) is essential to secure iris recognition systems. Recent iris PAD solutions achieved good performance by leveraging deep learning techniques. However, most results were reported under intra-database scenarios and it is unclear if such solutions can generalize well across databases and capture spectra. These PAD methods run the risk of overfitting because of the binary label supervision during the network training, which serves global information learning but weakens the capture of local discriminative features. This chapter presents a novel attention-based deep pixel-wise binary supervision (A-PBS) method. A-PBS utilizes pixel-wise supervision to capture the fine-grained pixel/patch-level cues and attention mechanism to guide the network to automatically find regions where most contribute to an accurate PAD decision. Extensive experiments are performed on six NIR and one visible-light iris databases to show the effectiveness and robustness of proposed A-PBS methods. We additionally conduct extensive experiments under intra-/cross-database and intra-/cross-spectrum for detailed analysis.
The results of our experiments indicates the generalizability of the A-PBS iris PAD approach.
}

\abstract{
Iris Presentation Attack Detection (PAD) is essential to secure iris recognition systems. 
Recent iris PAD solutions achieved good performance by leveraging deep learning techniques. However, most results were reported under intra-database scenarios and it is unclear if such solutions can generalize well across databases and capture spectra. These PAD methods run the risk of overfitting because of the binary label supervision during the network training, which serves global information learning but weakens the capture of local discriminative features. This chapter presents a novel attention-based deep pixel-wise binary supervision (A-PBS) method. A-PBS utilizes pixel-wise supervision to capture the fine-grained pixel/patch-level cues and attention mechanism to guide the network to automatically find regions where most contribute to an accurate PAD decision. Extensive experiments are performed on six NIR and one visible-light iris databases to show the effectiveness and robustness of proposed A-PBS methods. We additionally conduct extensive experiments under intra-/cross-database and intra-/cross-spectrum for detailed analysis.
The results of our experiments indicates the generalizability of the A-PBS iris PAD approach.
}

\section{Introduction} 
\label{sec:int}

Iris recognition systems are increasingly being deployed in many law enforcement and civil applications \cite{DBLP:journals/prl/JainNR16, DBLP:journals/ivc/Boutros20, DBLP:conf/ijcb/Boutros20}. 
However, iris recognition systems are vulnerable to Presentation Attacks (PAs) \cite{livedet17, livdet2020}, performing to obfuscate the identity or impersonate a specific person. Such attacks can be performed by various methods ranging from printouts, video replay, or textured contact lenses, among others. {As a result, the Presentation Attack Detection (PAD) domain has been developing solutions to mitigate security concerns in recognition systems.}

Recent iris PAD works \cite{crossdomain19, DBLP:conf/fusion/FangDBKK20, DBLP:conf/icb/FangDKK20, DBLP:conf/icb/SharmaR20, DBLP:journals/ivc/FangDBKK21, DBLP:conf/eusipco/FangDKK20} are competing to boost the performance using Convolution Neural Network (CNN) to facilitate discriminative feature learning. Even though the CNN-based algorithms achieved good results under intra-database setups, {
their generalizability across unseen attacks, databases, and spectra is still understudied. The results reported on the LivDet-Iris competitions verified the challenging nature of cross-PA and cross-database PAD.} The LivDet-Iris is an international competition series launched in 2013 to assess the current state-of-the-art in the iris PAD field. The two most recent edition took place in 2017 \cite{livedet17} and 2020 \cite{livdet2020}. The results reported in the LivDet-Iris 2017 \cite{livedet17} databases pointed out that there are still advancements to be made in the detection of iris PAs, especially under cross-PA, cross-sensor, or cross-database scenarios. Subsequently, LivDet-Iris 2020 \cite{livdet2020} reported a significant performance degradation on novel PAs, showing that the iris PAD is still a challenging task.
{Iris PAD in the NIR domain has so far shown good performances and indicated the generalizability challenges under cross-database scenarios \cite{DBLP:journals/ivc/FangDBKK21,crossdomain19,DBLP:journals/mva/FangDBKK22}. Nonetheless, studies addressing PAD algorithms in the visible spectrum are relatively limited \cite{DBLP:conf/btas/RajaRB15a,DBLP:journals/tifs/RaghavendraB15,DBLP:conf/icb/YadavKVSN17}. Given that the integration of iris recognition in smart devices is on the rise \cite{samsung_iris_scanner,payeye,DBLP:conf/btas/RajaRB15}, the study of iris PAD under the visible spectrum is essential. 
Furthermore, knowing that most iris PAD solutions are developed and trained for images captured in the NIR domain, an investigation of cross-spectrum iris PAD performance is much needed.
To our knowledge, there is no existing work on that investigated the PAD performance under a cross-spectrum scenario. As a result, in this work we further address the a visible-light-based iris PAD and the cross-spectrum PAD scenario.}

{Most of the recent iris PAD solutions trained models by binary supervision (more details in Sec.~\ref{sec:rw}), i.e., networks were only informed that an iris image is bona fide or attack, which may lead to overfitting. Besides, the limited binary information may be inefficient in locating the regions that contribute most to making an accurate decision. To target these issues, an Attention-based Pixel-wise Binary Supervision (A-PBS) network (See Figure \ref{fig:networks}) is proposed. In this chapter, we adopt the A-PBS solution to perform extensive experiments under intra- and cross-spectrum scenarios. The main contributions of the chapter include 1) We present the A-PBS solution that successfully aim to capture subtle and fine-grained local features in attack iris samples with the help of spatially positional supervision and attention mechanism. 2) We perform extensive experiments on NIR-based LivDet-Iris 2017 databases, three publicly available NIR-based databases, and one visible-spectrum-based iris PAD database. The experimental results indicated that the A-PBS solution outperforms state-of-the-art PAD methods in most experimental settings, including cross-PA, cross-sensor, and cross-databases. 3) We additionally analyze the cross-spectrum performance of the presented PAD solutions. To our best knowledge, this is the first work in which the cross-spectrum iris PAD performance is investigated.}

\section{Related Works}
\label{sec:rw}

\textbf{CNN-based iris PAD:} In recent years, many works \cite{DBLP:conf/icb/FangDKK20, DBLP:conf/fusion/FangDBKK20, DBLP:conf/icb/SharmaR20, crossdomain19, fusionvgg18, DBLP:conf/eusipco/FangDKK20} leveraged deep learning techniques and showed great progress in iris PAD performance. Kuehlkamp \textit{et al.} \cite{crossdomain19} {proposed to combine multiple} CNNs with the hand-crafted features. 
{Nevertheless, training $61$ CNNs requires} high computational resources and can be considered as an over-tailored solution. Yadav \textit{et al.} \cite{fusionvgg18} employed the fusion of hand-crafted features with CNN features and achieved good results.
Unlike fusing the hand-crafted and CNN-based features, Fang \textit{et al.} \cite{DBLP:conf/fusion/FangDBKK20} {presented a multi-layer deep features fusion approach (MLF) by considering} the characteristics of networks that different convolution layers encode the different levels of information. Apart from such fusion methods, a deep learning-based framework named Micro Stripe Analyses (MSA) \cite{DBLP:conf/icb/FangDKK20, DBLP:journals/ivc/FangDBKK21} was introduced to capture the artifacts around the iris/sclera boundary and showed a good performance on textured lens attacks. Yadav \textit{et al.} \cite{densepad19} presented DensePAD method to detect PAs by utilizing DenseNet architecture \cite{densenet}. Their experiments demonstrated the efficacy of DenseNet in the iris PAD task. 
Furthermore, Sharma and Ross \cite{DBLP:conf/icb/SharmaR20} exploited the architectural benefits of DenseNet \cite{densenet} to propose an iris PA detector (D-NetPAD) evaluated on a proprietary database and the LivDet-Iris 2017 databases. 
{With the help of their private additional data, the fine-tuned D-NetPAD achieved good results on LivDet-Iris 2017 databases, however, scratch D-NetPAD failed in the case of cross-database scenarios.} 
These works inspired us to use DenseNet \cite{densenet} as the backbone for our {A-PBS network} architectures.
Recently, Chen \textit{et al.} \cite{DBLP:conf/wacv/ChenR21} proposed an attention-guided iris PAD method to refine the feature maps of DenseNet \cite{densenet}. {However, this method utilized conventional sample binary supervision and did not report cross-database and cross-spectrum experiments to prove the generalizability of the additional attention module. }

\textbf{Limitations:} 
Based on the recent iris PAD literature, it can be concluded that deep-learning-based methods boost the performance but still {have the risk of overfitting under cross-PA, cross-database, and cross-spectrum scenarios. Some} recent methods proposed the fusion of multiple PAD systems or features to improve the generalizability \cite{DBLP:conf/fusion/FangDBKK20,crossdomain19,fusionvgg18}, which makes it challenging for deployment. One of the major reasons causing overfitting is the lack of availability of a sufficient amount of variant iris data for training networks. Another possible reason might be binary supervision. While the binary classification model provides useful global information, its ability to capture subtle differences in attacking iris samples may be weakened, and thus the deep features might be less discriminative. This possible cause motivates us to exploit binary masks to supervise the training of our PAD model, because a binary mask label may help to supervise the information at each spatial location. However, PBS may also lead to another issue, as the model misses the exploration of important regions due to the 'equal' focus on each pixel/patch. To overcome some of these difficulties, we propose the A-PBS architecture to force the network to find regions that should be emphasized or suppressed for a more accurate iris PAD decision. The detailed introduction of PBS and A-PBS can be found in Sec.~\ref{sec:meh}.
{In addition to the challenges across database scenarios, another issue is that there is no existing research dedicated to exploring the generalizability of PAD methods across spectral scenarios. PAD research in the NIR domain \cite{fusionvgg18,crossdomain19,DBLP:conf/icb/SharmaR20,DBLP:conf/fusion/FangDBKK20,DBLP:conf/icb/FangDKK20} has attracted much attention, while few studies \cite{DBLP:conf/btas/RajaRB15a,DBLP:journals/tifs/RaghavendraB15,DBLP:conf/icb/YadavKVSN17} investigated PAD performance in the visible spectrum. Furthermore, the generalizability of PAD methods under the cross-spectrum is unclear. Suppose a model trained on NIR data can be well generalized to visible-light data. In that case, it requires only low effort to transfer such solution to low-cost mobile devices \cite{samsung_iris_scanner,payeye,DBLP:conf/btas/RajaRB15}, which simplifies its application in the real world. Therefore, this chapter explores the proposed PAD performance in a cross-spectrum experimental settings.}

\section{Methodology} 
\label{sec:meh}

\begin{figure}[htbp!]
\caption{An overview of (a) baseline DenseNet, (b) proposed PBS and (c) proposed A-PBS networks. \label{fig:networks}}
\centering{\includegraphics[width=0.99\linewidth]{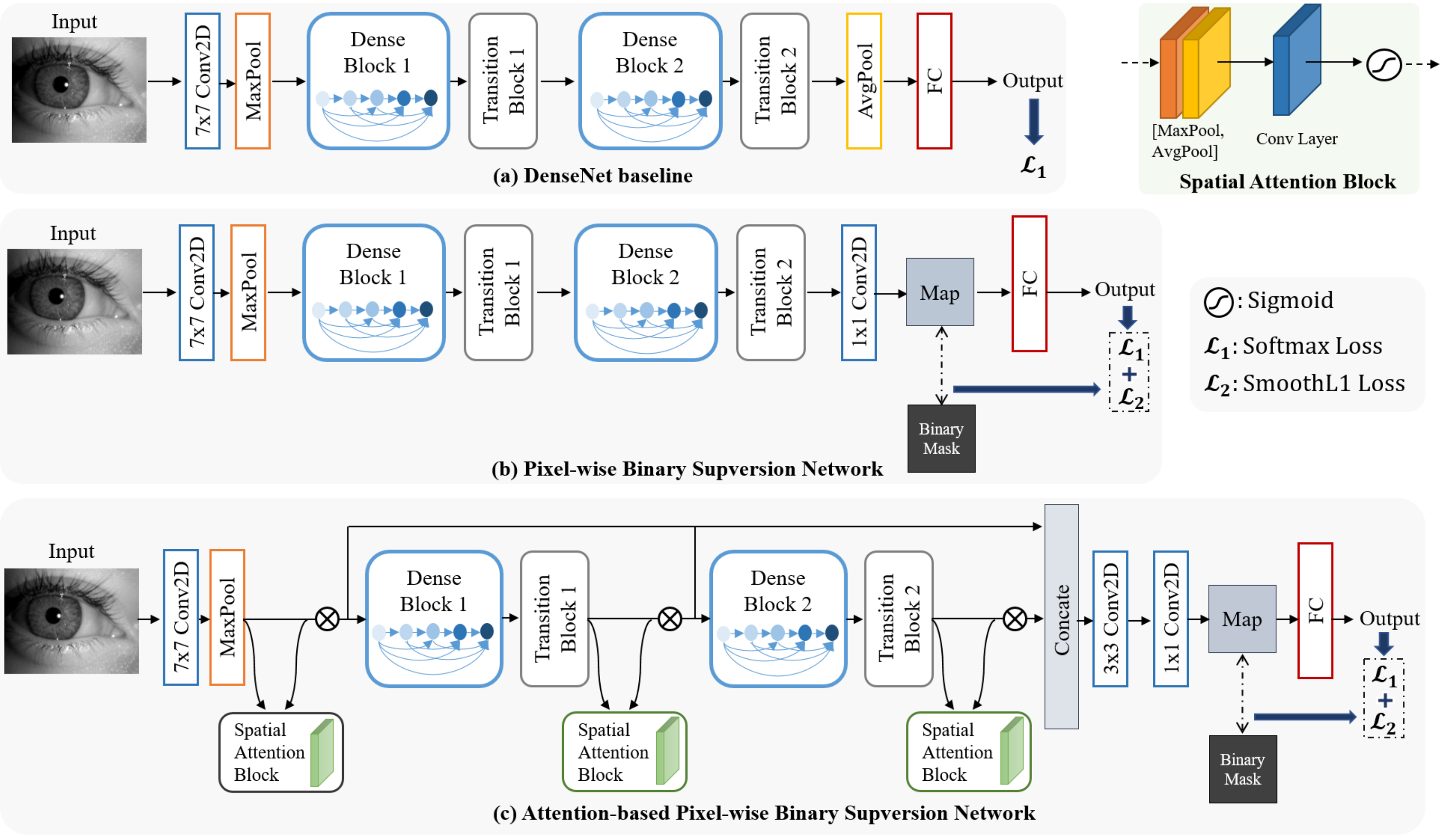}}
\end{figure}

{This section starts by introducing the DenseNet \cite{densenet}, which is used as a preliminary backbone architecture. Then, the Pixel-wise Binary Supervision (PBS) and Attention-based PBS (A-PBS) methods are described. We presented this approach initially in \cite{DBLP:conf/icb/FangDBKK21}, however, we extend it here by investigating its advantages on different attack types, iris images captured in the visible spectrum, and cross-spectrum deployments. Figure \ref{fig:networks} depicts an overview of our different methods. The first gray block (a) presents the basic DenseNet architecture with binary supervision,} the second gray block (b) introduces the binary and PBS, and the third block (c) is the PBS with the fused multi-scale spatial attention mechanism (A-PBS).

\subsection{Baseline: DenseNet}
DenseNet \cite{densenet} presented direct connection between any two layers with the same feature-map size in a feed-forward fashion. 
{The reasons inspiring our selection of DensetNet are: 1) DenseNets naturally integrate the properties of identity mappings and deep supervision following a simple connectivity rule. 2) DenseNet has already demonstrated its superiority in iris PAD \cite{densepad19, DBLP:conf/icb/SharmaR20, livdet2020}. Figure~\ref{fig:networks}.(a) illustrates that we reuse two dense and transition blocks of pre-trained DenseNet121.} An average pooling layer and a fully-connected (FC) classification layer are sequentially appended, following the second transition block, to generate the final prediction to determine whether the iris image is bona fide or attack. PBS and A-PBS networks are extended on this basic architecture later.

\subsection{Pixel-wise Binary Supervision Network (PBS)}
{By reviewing the recent} iris PAD literature \cite{DBLP:conf/icb/FangDKK20, DBLP:conf/fusion/FangDBKK20, DBLP:conf/icb/SharmaR20, crossdomain19}, it can be found that CNN-based methods outperformed hand-crafted feature-based methods. In typical CNN-based iris PAD solutions, networks are designed such that feeding pre-processed iris image as input to learn discriminative features between bona fide and artifacts. To that end, a FC layer is generally introduced to output a prediction score supervised by binary label (bona fide or attack). {Recent face PAD works have shown that auxiliary supervision \cite{DBLP:conf/cvpr/LiuJ018, deeppixbis, DBLP:journals/wacv22/Fang22} achieved significant improvement in detection performance}. Binary label supervised classification learns semantic features by capture global information but may lead to overfitting. Moreover, such embedded 'globally' features might lose the local detailed information in spatial position. These drawbacks give us the insight that {adding pixel-wise binary along with binary supervision might improve the PAD performance}. First, such supervision approach can be seen as a combination of patch-based and vanilla CNN based methods. To be specific, each pixel-wise score in output feature map is considered as the score generated from the patches in an iris image. Second, the binary mask supervision would be provided for the deep embedding features in each spatial position. 
{Figure~\ref{fig:networks}.(b) illustrates the network details that an intermediate feature map is produced before the final binary classification layer. The output from the \textit{Transition Block 2} is 384 channels with the map size of $14 \times 14$. A $1 \times 1$ convolution layer is added to produce the intermediate map. In the end, an FC layer is utilized to generate a prediction score.}

\subsection{Attention-based PBS Network (A-PBS)}
The architecture of PBS is designed coarsely (simply utilizing the intermediate feature map) based on the DenseNet \cite{densenet}, which might be sub-optimal for iris PAD task. To enhance that, and inspired by Convolutional Block Attention Mechanism (CBAM) \cite{cbam} and MLF \cite{DBLP:conf/fusion/FangDBKK20}, we propose an A-PBS method with multi-scale feature fusion (as shown in Figure~\ref{fig:networks}.(c)). 

{Even though PBS boosts iris PAD performance under intra-database/-spectrum, it shows imperfect invariation under more complicated cross-PA, cross-database, and cross-spectrum scenarios (See results in Table~\ref{tab:cross_db}, \ref{tab:nir_vis}, and \ref{tab:vis_nir}).} As a result, it is worth finding the important regions to focus on, although it contradicts learning \textit{more} discriminative features. In contrast, the attention mechanism aims to automatically learn \textit{essential} discriminate features from inputs that are relevant to PA detection. Woo \textit{et al.} \cite{cbam} presented an attention module consisting of the channel and spatial distinctive sub-modules, {which possessed consistent improvements in various classification and detection tasks across different network architectures. Nonetheless,} only spatial attention module is employed in our case due to the following reasons. {The first reason is that} the Squeeze-and-Excitation (SE) based channel attention module focuses only on the inter-channel relationship by using dedicated global feature descriptors. Such channel attention module may lead to a loss of information (e.g., class-deterministic pixels) and may result in further performance degradation when the domain is shifted, e.g., different sensors and changing illumination. {Second, a benefit of the spatial attention module is that the inter-spatial relationship of features is utilized.} Specifically, it focuses on \textit{'where'} is an informative region, which is more proper for producing intermediate feature maps for supervision. Furthermore, based on the fact that the network embeds different layers of information at different levels of abstraction, the MLF \cite{DBLP:conf/fusion/FangDBKK20} approach confirmed that the fusing deep feature from multiple layers is beneficial to enhance the robustness of the networks in the iris PAD task. Nevertheless, we propose to fuse feature maps generated from different levels directly within the network instead of fusing features extracted from a trained model in MLF \cite{DBLP:conf/fusion/FangDBKK20}, {because finding} the best combination of network layers to fuse is a challenging task and difficult to generalize well, especially when targeting different network architectures.

{Figure~\ref{fig:networks} illustrates that three spatial attention modules are appended after \textit{MaxPool}, \textit{Transition Block 1}, and \textit{Transition Block 2}, respectively.} The feature learned from the \textit{MaxPool} or two \textit{Transition Blocks} can be considered as low-, middle- and high-level features and denoted as 
\begin{equation}
\mathcal{F}_{level} \in \mathbb{R}^{C \times H \times W}, \quad {level} \in \{low, mid, high\}\;.
\end{equation}
Then, the generated attention maps $\mathcal{A}_{level} \in \mathbb{R}^{H \times W} $ encoding where to emphasize or suppress are used to refine $\mathcal{F}_{level}$. The refined feature $\mathcal{F}'_{level}$ can be formulated as $\mathcal{F}'_{level} = \mathcal{F}_{level} \otimes \mathcal{A}_{level}$ where $\otimes$ is matrix multiplication. Finally, such three different level refined features are concatenated together and then fed into a $1 \times 1$ convolution layer to produce the pixel-wise feature map for supervision. It should be noticed that the size of convolutional kernel in three spatial attention modules is different. As mentioned earlier, the deeper the network layer, the more complex and abstract the extracted features. Therefore, we should use smaller convolutional kernels for deeper features to locate useful region. The kernel sizes of low-, middle- and high-level layers are thus set to 7, 5, and 3, respectively. The experiments have been demonstrated later in Sec.~\ref{sec:es} and showed that {in most experimental setups, the A-PBS solution exhibited superior performance and generalizability in comparison to the PBS and DenseNet approaches.}

\subsection{Loss Function}
{In the training phase,} Binary Cross Entropy (BCE) loss is used for final binary supervision. For the sake of robust PBS needed in iris PAD, Smooth L1 (SmoothL1) loss is utilized to help the network reduce its sensitivity to outliers in the feature map. The equations for SmoothL1 is shown below:
\begin{equation}
\mathcal{L}_{SmoothL1} = \dfrac{1}{n} \sum z \;,
where \quad z = 
\begin{cases}
\dfrac{1}{2}\cdot (y - x)^{2}, & if \quad |y - x| < 1 \\
|y-x| - \dfrac{1}{2}, & {otherwise}
\end{cases}
\end{equation}
$n$ is the amount number of pixels in the output map ($14 \times 14$ in our case). The equation of BCE is:
\begin{equation}
\mathcal{L}_{BCE} = -[y \cdot \log p + (1-y)\cdot \log (1-p)]\;,
\end{equation}
where $y$ in both loss equations presents the ground truth label. $x$ in SmoothL1 loss presents to the value in feature map, while $p$ in BCE loss is predicted probability.
The overall loss $\mathcal{L}_{overall}$ is formulated as $\mathcal{L}_{overall} = \lambda \cdot \mathcal{L}_{SmoothL1} + (1 - \lambda) \cdot \mathcal{L}_{BCE}$. {In our experiments, the $\lambda$ is set to 0.2.}

\subsection{Implementation Details}
{In the training phase, we performed class balancing by under-sampling the majority class for the databases, whose distribution of bona fides and attacks are imbalanced in the training set.} Data augmentation was performed during training using random horizontal flips with a probability of 0.5. The model weight of DenseNet, PBS and A-PBS models were first initialized by the base architecture DenseNet121 trained on the ImageNet dataset and then fine-tuned by iris PAD data, by considering the limited amount of iris data. The Adam optimizer was used for training with a initial learning rate of $1e^{-4}$ and a weight decay of $1e^{-6}$. To further avoid overfitting, the model was trained with the maximum 20 epochs and the learning rate halved every 6 epochs. The batch size is 64. In the testing stage, {the binary output was used as a final prediction score}. The proposed method was implemented using the Pytorch.
\section{Experimental Evaluation}
\label{sec:es}

\subsection{Databases}
\label{ssec:db}
\begin{figure}[htbp!]
\centering{\includegraphics[width=1.0\linewidth]{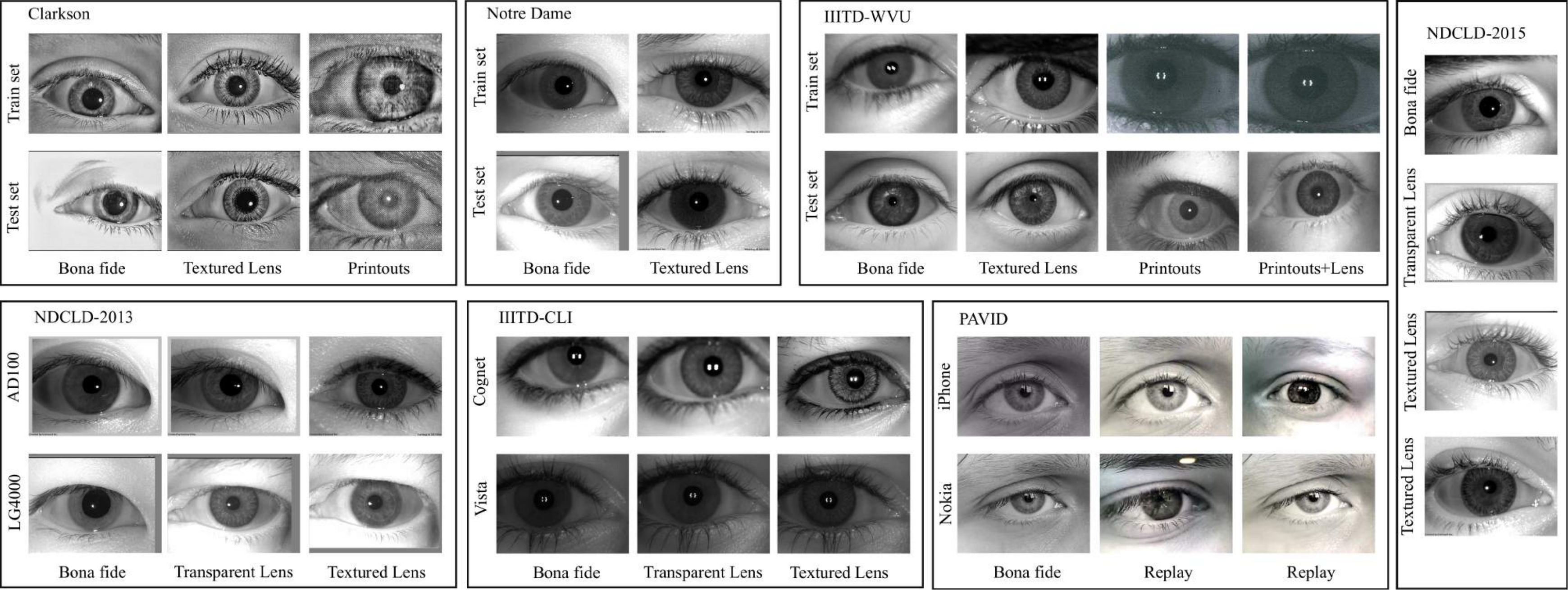}}
\caption{{Iris image samples from the used databases. It should be noted that transparent lens is classified as bona fide in our case. Only PAVID database was captured under the visible spectrum.}}
\label{fig:iris_samples}
\end{figure}

{The DenseNet, PBS, and A-PBS were evaluated on multiple databases: three NIR-based databases comprising of textured contact lens attacks captured by different sensors \cite{ndcld15, ndcld2013, iiitd_cli}, and three databases (Clarkson, Notre Dame and IIITD-WVU) from the LivDet-Iris 2017 competition \cite{livedet17} (also NIR-based).} The Warsaw database in the LivDet-Iris 2017 is no longer publicly available due to General Data Protection Regulation (GDPR) issues. For the experiments on NDCLD13, NDCLD15, IIIT-CLI databases, 5-fold cross-validation was performed due to no pre-defined training and testing sets. For the experiments in competition databases, we followed the defined data partition and experimental setting \cite{livedet17}. {In addition to above NIR-based iris databases, we also perform experiments on another publicly available database where images were captured under the visible spectrum, named Presentation Attack Video Iris Database (PAVID) \cite{DBLP:conf/btas/RajaRB15a}.}
Subjects in each fold or defined partition are dis-joint. The image samples can be found in Figure \ref{fig:iris_samples} and the summery of the used databases is listed in Tab~\ref{Tab:db_description}. 

\begin{table}[thbp!]
\caption{Characteristics of the used databases. All databases have the training and test sets based on their own experimental setting in the related papers. The Warsaw database in Iris-LivDet-2017 competition are no longer publicly available. }
\label{Tab:db_description}
\begin{center}
\resizebox{1.0\textwidth}{!}{%
\begin{tabular}{l|l|l|l|l|l}
\hline
\multicolumn{2}{l|}{Database} & Spectrum & \# Training & \# Testing & Type of Iris Images   \\
\hline
\multicolumn{2}{l|}{NDCLD-2015 \cite{ndcld15}} & NIR & 6,000 & 1,300  & BF, soft and textured lens  \\ \hline
{NDCLD-2013 \cite{ndcld2013}} & LG4000  & {NIR} & 3,000 & 1,200  & BF, soft and textured lens   \\ 
& AD100 & NIR & 600 & 300 & BF, soft and textured lens   \\ \hline
{IIIT-D CLI \cite{iiitd_cli_2, iiitd_cli}} & Cognet & {NIR} & 1,723 & 1,785 & BF, soft and textured lens   \\ 
& Vista & NIR & 1,523 & 1,553 & BF, soft and textured lens    \\ \hline
{LivDet-Iris 2017 \cite{livedet17}} & Clarkson (cross-PAD) & {NIR} &  4937 & 3158 & BF, textured lens, printouts \\ 
& Notre Dame (cross-PA) & NIR & 1,200 & 3,600 &  BF, textured lenses  \\ 
& IIITD-WVU (cross-DB) & NIR& 6,250 & 4,209 & BF, textured lenses, printouts, lens printouts \\ \hline
\multicolumn{2}{l|}{PAVID \cite{DBLP:conf/btas/RajaRB15a}} & VIS & 180 $^a$ & 612 $^a$  & BF, replay  \\ 
\hline
\end{tabular}}
\end{center}
BF: bona fide, VIS: visible light, NIR: Near-Infrared light \\
$^a$ the format of data is video, others are images

\end{table}

\textbf{NDCLD-2013:} The NDCLD-2013 database {comprises of $5100$ NIR images and is conceptually divided into two sets based on capture sensors}: 1) LG4000 including $4200$ images captured by IrisAccess LG4000 camera, 2) AD100 consisting of $900$ images captured by risGuard AD100 camera. Both the training and the test set are divided equally into no lens (bona fide), soft lens (bona fide), and textured lens (attack) classes.

\textbf{NDCLD-2015:} The $7300$ images in the NDCLD-2015 \cite{ndcld15} were captured by two sensors, IrisGuard AD100 and IrisAccess LG4000 under MIR illumination and controlled environments. The NDCLD15 contains iris images wearing no lenses, soft lenses, textured lenses.

\textbf{IIIT-D CLI:} IIIT-D CLI database contains $6570$ iris images of $101$ subjects with left and right eyes. For each identity, three types of images were captured: 1) no lens, 2) soft lens, and 3) textured lens. Iris images are divided into two sets based on captured sensors: 1) Cogent dual iris sensor and 2) VistaFA2E single iris sensor.

\textbf{LivDet-Iris 2017 Database:} Though the new edition LivDet-Iris competition was held in 2020, we still evaluate the algorithms in databases provided by LivDet-Iris 2017 {for several reasons: 1) No official training data was announced in the LivDet-Iris 2020 because the organizers encouraged the participants to use all available data (both publicly and proprietary) to enhance the effectiveness and robustness. 2) The test data is not publicly available. Consequently, to make a fair comparison with state-of-the-art algorithms on equivalent data, we use LivDet-Iris 2017 databases to restrict the evaluation factors to the algorithm itself rather than the data. 3) The LivDet-Iris 2017 competition databases are still valuable due to the challenging cross-PA and cross-database scenario settings. }
The Clarkson and Notre Dame database are designed for cross-PA scenarios, while the IIIT-WVU database is designed for a cross-database evaluation due to the different sensors and acquisition environments. The Clarkson testing set includes additional unknown visible-light image printouts and unknown textured lenses (unknown pattern). Moreover, Notre Dame focused on the unknown textured lenses. However, the Warsaw database is no longer publicly available. 

{\textbf{Presentation Attack Video Iris Database (PAVID) \cite{DBLP:conf/btas/RajaRB15a}:} PAVID is the video iris database collected using smartphones (Nokia Lumia 1020 and iPhone 5S) in the visible spectrum. PAVID contains 304 bona fide videos and 608 replay attack videos across 76 subjects. Moreover, PAVID was divided into three sets in the official protocol: training set including 180 videos, development set including 120 videos, and testing set including 608 videos. The development set defined in \cite{DBLP:conf/btas/RajaRB15a} was used only for determining the filter kernel of the Laplacian pyramid in \cite{DBLP:conf/btas/RajaRB15a}, not for computing the decision threshold. Therefore, we omit the development set in our experiments. }

\subsection{Evaluation Metrics}

The following metrics are used to measure the PAD algorithm performance: 1) Attack Presentation Classification Error Rate (APCER), the proportion of attack images incorrectly classified as bona fide samples, 2) Bona fide Presentation Classification Error Rate (BPCER), the proportion of bona fide images incorrectly classified as attack samples, 3) Half Total Error Rate (HTER), the average of APCER and BPCER. The APCER and BPCER follows the standard definition presented in the ISO/IEC 30107-3 \cite{ISO301073} and are adopted in most PAD literature including in LivDet-Iris 2017. The threshold for determining the APCER and BPCER is 0.5 as defined in the LivDet-Iris 2017 protocol. In addition, {for further comparison with the state-of-the-art iris PAD algorithms on IIITD-CLI \cite{iiitd_cli, iiitd_cli_2} database}, we also report the Correct Classification Accuracy (CCR). CCR is the ratio between the total number of correctly classified images and the number of all classified presentations. {Furthermore, to enable the direct comparison with \cite{DBLP:conf/icb/SharmaR20}, we evaluate the performance of our presented DenseNet, PBS, and A-PBS methods} in terms of True Detection Rate (TDR) at a false detection rate of 0.2\%, as \cite{DBLP:conf/icb/SharmaR20} claims that this threshold is normally used to demonstrate the PAD performance in practice. 
TDR is 1 -APCER, and false detection rate is defined to be the same as BPCER, we therefore use BPCER. An Equal Error Rate (EER) locating at the intersection of APCER and BPCER {is also reported under cross-database and cross-spectrum settings (results as shown in Table~\ref{tab:cross_db}, \ref{tab:vis_nir}, and \ref{tab:nir_vis}).} The metrics beyond APCER and BPCER are presented to enable a direct comparison with reported results in state-of-the-arts.  

{\section{Intra-spectrum and cross-database evaluation results}}
{This section presents the evaluation results on different databases and comparison to state-of-the-art algorithms. The comparison to state-of-the-arts depends mainly on the reported results in the literature, as most algorithms are not publicly available or their technical description is insufficient to ensure error-free re-implementation. Therefore, we aim to report the widest range of metrics used in other works to enable an extensive comparison. First, the results from different aspects/metrics on LivDet-Iris 2017 database are reported in Table \ref{tab:livdet17_results} which compare our solution with state-of-the-art PAD methods, \ref{tab:livdet17_tdr} that report the results in terms of TDR at low BPCER, and \ref{tab:PA_errors} that investigate the performance on different PAs. Then, we demonstrate the experiments under cross-database scenarios by using the three databases in LivDet-Iris 2017 competition to verify the generalizability of our A-PBS solution. Furthermore, the results on NDCLD-2013/NDCLD-2015 and IIITD-CLI databases are presented in Table \ref{tab:cld_results}, \ref{tab:tdr_results_nd_iiit} and Table \ref{tab:iiit_cld_results}, respectively. We further perform the experiment on the PAVID database in visible spectrum (results in Table \ref{tab:intra-pavid}). In this section, we also provide explainability analyses using attention map visualisations for further visual reasoning of the presented solution. }

\subsection{Iris PAD in the NIR spectrum}

\subsubsection{Results on the LivDet-Iris 2017 Databases}
Table~\ref{tab:livdet17_results} summarizes the results in terms of APCER, BPCER, and HTER on the LivDet-Iris 2017 databases. 
We evaluate the algorithms on databases provided by LivDet-Iris 2017. The evaluation and comparison on LivDet-Iris 2020 are not included due to 1) no officially offered training data, 2) not publicly available test data. Moreover, {LivDet-Iris 2017 databases are still considered as a challenging task, because the experimental protocols are designed for complicated cross-PA and cross-database scenarios.} {In this chapter, we aim to focus on the impact of the algorithm itself on PAD performance rather than the diversity of data.} Consequently, to make a fair comparison with state-of-the-art algorithms on equivalent data, we compare to the Scratch version of the D-NetPAD results \cite{DBLP:conf/icb/SharmaR20}, because Pre-trained and Fine-tuned D-NetPAD used additional data (including part of Notre Dame test data) for training. This was not an issue with the other compared state-of-the-art methods.

\begin{table}[thbp!]
\caption{Iris PAD performance of {our presented DenseNet, PBS, and A-PBS solutions}, and existing state-of-the-art algorithms on LivDet-Iris 2017 databases in terms of APCER (\%), BPCER (\%) and HTER (\%) which determined by a threshold of 0.5. The \textit{Winner} in first column refers to the winner of each competition database. Bold numbers indicate the two lowest HTERs.}
\label{tab:livdet17_results}
\resizebox{1.0\textwidth}{!}{%
\begin{tabular}{c|c|c|c|c|c|c|c||c|c|c}
\hline
{Database} & {Metric} & Winner \cite{livedet17} & SpoofNet \cite{spoofnet_tuning} & Meta-Fusion \cite{crossdomain19} & D-NetPAD \cite{DBLP:conf/icb/SharmaR20} & MLF \cite{DBLP:conf/fusion/FangDBKK20} & MSA \cite{DBLP:conf/icb/FangDKK20, DBLP:journals/ivc/FangDBKK21} & DenseNet & PBS & A-PBS \\
\hline
\multirow{3}{*}{Clarkson} & APCER & 13.39 & 33.00 & 18.66 & 5.78 & - & - & 10.64 & 8.97 & 6.16 \\ 
 & BPCER & 0.81 & 0.00 & 0.24 & 0.94 & - & - & 0.00 & 0.00 & 0.81 \\ 
 & HTER & 7.10 & 16.50 & 9.45 & \textbf{3.36} & - & - & 5.32 & 4.48 & \textbf{3.48} \\
\hline
\multirow{3}{*}{Notre Dame} & APCER & 7.78 & 18.05 & 4.61 & 10.38 & 2.71 & 12.28 & 16.00 & 8.89 & 7.88 \\
 & BPCER & 0.28 & 0.94 & 1.94 & 3.32 & 1.89 & 0.17 & 0.28 & 1.06 & 0.00 \\ 
 & HTER & 4.03 & 9.50 & \textbf{3.28} & 6.81 & \textbf{2.31} & 6.23 & 8.14 & 4.97 & 3.94 \\ 
\hline
\multirow{3}{*}{IIITD-WVU} & APCER & 29.40 & 0.34 & 12.32 & 36.41 & 5.39 & 2.31 & 2.88 & 5.76 & 8.86 \\
 & BPCER & 3.99 & 36.89 & 17.52 & 10.12 & 24.79 & 19.94 & 17.95 & 8.26 & 4.13 \\
 & HTER & 16.70 & 18.62 & 14.92 & 23.27 & 15.09 & 11.13 & 10.41 & \textbf{7.01} & \textbf{6.50} \\
\hline
\end{tabular}}
\end{table}

\begin{table}[thbp!]
\caption{Iris PAD performance reported in terms of TDR (\%) at 0.2\% BPCER on the LivDet-Iris 2017 databases. K indicates known test subset and U is unknown subset. The highest TDR is in bold.}
\label{tab:livdet17_tdr}
\begin{center}
\begin{tabular}{c|c|c|c|c|c}
\hline
\multicolumn{2}{c|}{\multirow{2}{*}{Database}} & \multicolumn{4}{c}{TDR (\%) @ 0.2\% BPCER} \\ \cline{3-6} 
\multicolumn{2}{c|}{} & D-NetPAD \cite{DBLP:conf/icb/SharmaR20} & DenseNet & PBS & A-PBS \\ \hline
\multicolumn{2}{c|}{Clarkson} & 92.05 & 92.89 & \textbf{94.02} & 92.35 \\ \hline
\multirow{2}{*}{Notre Dame} & K & \textbf{100.00} & 99.68 & 99.78 & 99.78 \\ 
 & U & 66.55 & 58.33 & 76.89 & \textbf{90.00} \\ \hline
\multicolumn{2}{c|}{IIITD-WVU} & 29.30 & 58.97 & 69.32 & \textbf{72.00} \\ \hline
\end{tabular}
\end{center}
\end{table}

\begin{figure}[htbp!]
\centering{\includegraphics[width=0.99\linewidth]{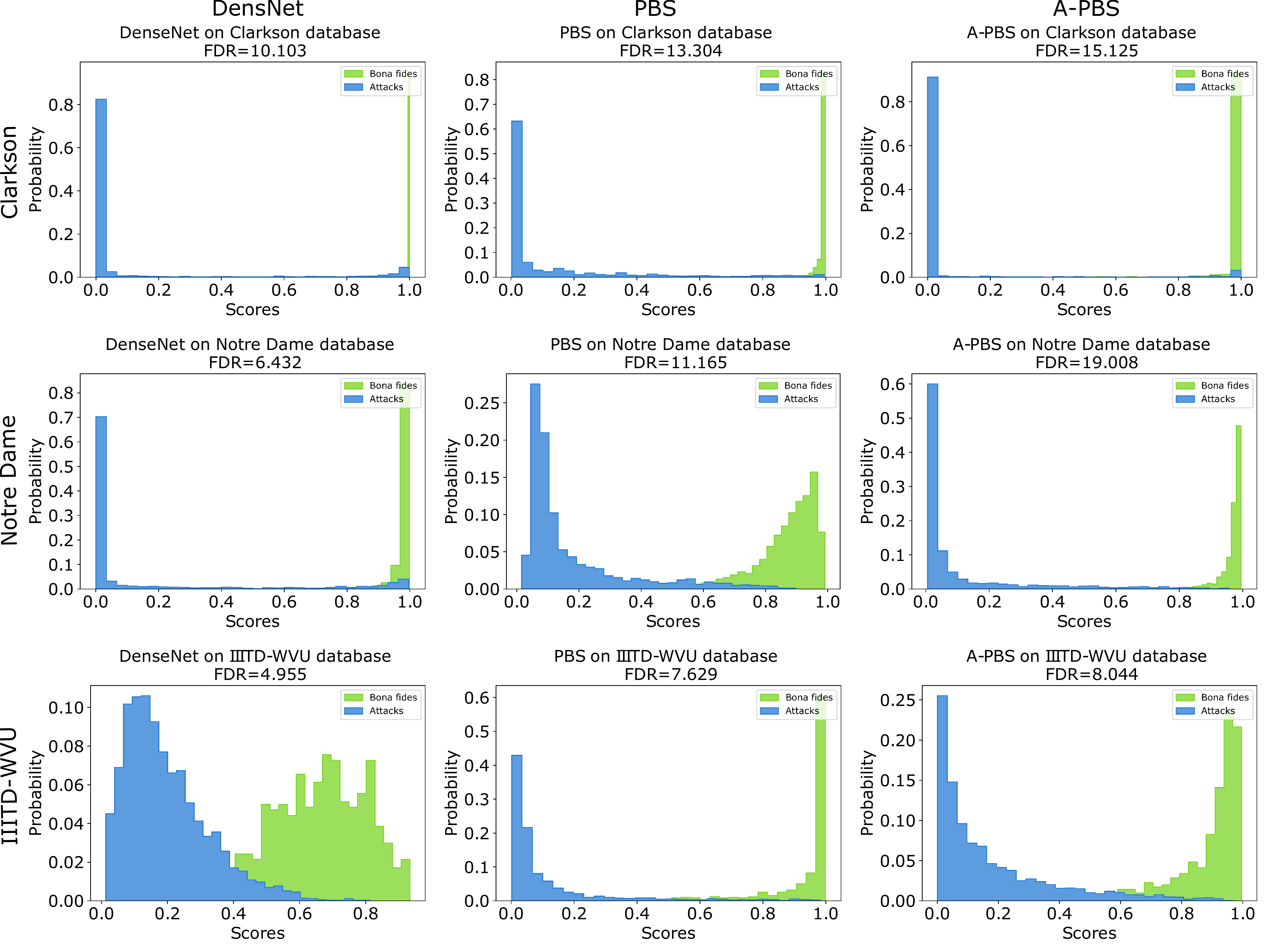}}
\caption{{PAD score distribution of bona fide (green) and PAs (blue) on the LivDet-Iris 2017 databases. The histogram top to bottom are results on Clarkson, Notre Dame and IIITD-WVU databases, and the histograms from left to right are produced by DenseNet, PBS, and A-PBS, respectively. The larger separability (measured by Fisher Discriminant Ratio (FDR) \cite{DBLP:journals/ijon/LorenaC10,DBLP:conf/eusipco/DamerON14}) and smaller overlap indicate higher classification performance. It can be observed that the our proposed A-PBS method achieved the highest FDR value on all three databases.}}
\label{fig:score_distribution}
\end{figure}

It can be observed in Table~\ref{tab:livdet17_results} that A-PBS architecture achieves significantly improved performance in comparison to DenseNet and also slightly lower HTER values than the PBS model in all cases. For instance, the HTER value on Notre Dame is decreased from 8.14\% by DenseNet and 4.97\% by PBS to 3.94\% by A-PBS. Although the slightly worse results on Notre Dame might be caused by the insufficient data in the training set, our PBS and A-PBS methods show significant superiority on the most challenging IIITD-WVU database. {Moreover, Figure \ref{fig:score_distribution} illustrates the PAD score distribution of the bona fide and PAs for further analysis.} The PAD score distribution generated by A-PBS shows an evident better separation between bona fide (green) and PAs (blue). In addition to reporting the results determined by a threshold of 0.5, we also measure the performance of DenseNet, PBS, and A-PBS in terms of its TDR at 0.2\% BPCER (to follow state-of-the-art trends \cite{DBLP:conf/icb/SharmaR20}) in Table~\ref{tab:livdet17_tdr}. It is worth noting that our A-PBS method achieves the highest TDR value (90.00\%) on unknown-test set in Notre Dame, while the second-highest TDR is 76.89\% achieved by PBS. 

\begin{table}[thbp!]
\caption{{Iris PAD performance reported based on each presentation attack on the LivDet-Iris-2017 database in terms of BPCER (\%) and APCER (\%). The Notre Dame database is omitted because it comprises only texture contact lens attack and the results are the same as in Table \ref{tab:livdet17_results}. It can be observed that textured contact lens attack is more challenging than printouts in most cases.}}
\label{tab:PA_errors}
\begin{center}
\resizebox{1.0\textwidth}{!}{%
\begin{tabular}{c|ccc|cccc}
\hline
Database & \multicolumn{3}{c|}{Clarkson} & \multicolumn{4}{c}{IIITD-WVU} \\
\hline
$\#$ Images  & $\#$ 1485 & $\#$ 908      & $\#$ 765      & $\#$ 704 & $\#$ 1404 & $\#$ 701 & $\#$ 1402 \\ \hline
Metric  & BPCER  & APCER (PR) & APCER (CL) & BPCER & APCER (PR) & APCER (CL) & APCER (PR-CL) \\ \hline
DenseNet & 0.00 & 0.66 & 22.48 & 17.95 & 3.06 & 8.27 & 0.00  \\ \hline
PBS   & 0.00 & 0.44  & 19.08  & 8.26  & 11.68 & 5.42 & 0.00 \\ \hline
A-PBS & 0.81 & 1.32& 10.59 & 4.13 & 11.68 & 17.97      & 0.86  \\ \hline
\end{tabular}}
\end{center}
PR: printouts, CL: textured contact lens, PR-CL: printed textured contact lens
\end{table}

{Furthermore, we explore the PAD performance based on each presentation attack in LivDet-Iris 2017 database \cite{livedet17}. Because the Notre Dame database contains only textured contact lenses, we report the results on Clarkson and IIITD-WVU databases in Table \ref{tab:PA_errors}. The results show that textured contact lens attacks obtain higher APCER values than printouts attack in most cases, e.g., the APCER value on textured lens attack is 10.59\% and on printouts is 1.52\% both achieved by A-PBS solution. Hence, we conclude that contact lens is more challenging than printouts in most cases.}

{In addition to intra-dataset evaluation, we }further evaluate the generalizability of our models under cross-database scenario, e.g., the model trained on Notre Dame is tested on Clarkson and IIITD-WVU. As shown in Table~\ref{tab:cross_db}, the A-PBS model outperforms DenseNet and PBS in most cases, which verifying that additional spatial attention modules can reduce the overfitting of the PBS model and capture fine-grained features. Furthermore, the DenseNet and A-PBS models trained on Notre Dame even exceed the prior state-of-the-arts when testing on the IIIT-WVU database (8.81\% HTER by DenseNet and 8.95\% by A-PBS, while the best prior state-of-the-art achieved 11.13\% (see Table~\ref{tab:livdet17_results})). It should be noted that the {APCER values on Notre Dame are significant higher by using models either trained on Clarkson or IIITD-WVU. Because Notre Dame training dataset contains only textured lens attacks while Clarkson and IIIT-WVU testing datasets comprise of both textured lens and printouts attacks, which makes this evaluation scenario partially consider unknown PAs.} In such an unknown-PAs situation, our A-PBS method achieved significantly improved results. In general, the cross-database scenario is still a challenging problem since many D-EER values are above 20\% (Table~\ref{tab:cross_db}).

\begin{table}[]
\caption{Iris PAD performance measured under cross-database scenarios and reported in terms of EER (\%), HTER (\%), APCER (\%), and BPCER (\%). APCER and BPCER are determined by a threshold of 0.5. The lowest error rate is in bold.}
\begin{center}
\begin{tabular}{c|cccc|cccc}
\hline
Train dataset & \multicolumn{8}{c}{Notre Dame} \\ \hline
Test dataset  & \multicolumn{4}{c|}{Clarkson}   & \multicolumn{4}{c}{IIITD-WVU} \\ \hline
Metric        & EER    & HTER  & APCER & BPCER & EER   & HTER  & APCER & BPCER \\ \hline
DenseNet      & 30.43  & 32.01 & 51.29 & \textbf{12.73} & 7.84  & \textbf{8.81}  & \textbf{5.93}  & 11.69 \\ 
PBS           & 48.36  & 47.28 & 28.15 & 66.4  & 15.52 & 14.54 & 22.24 & 6.83  \\ 
APBS          & \textbf{20.55}  & \textbf{23.24} & \textbf{14.76} & 31.72 & \textbf{6.99}  & 8.95  & 15.34 & \textbf{2.56}  \\ \svhline 
Train dataset & \multicolumn{8}{c}{Clarkson} \\ \hline
Test dataset  & \multicolumn{4}{c|}{Notre Dame} & \multicolumn{4}{c}{IIITD-WVU} \\ \hline
Metric        & EER    & HTER  & APCER & BPCER & EER   & HTER  & APCER & BPCER \\ \hline
DenseNet      & 22.33  & 31.11 & 62.22 & 0.00  & 26.78 & 42.40  & 84.80 & 0.00     \\ 
PBS           & 28.61  & 32.42 & 64.83 & 0.00  & 25.78 & 42.48 & 84.97 & 0.00     \\ 
APBS          & \textbf{21.33}  & \textbf{23.08} & \textbf{46.16} & 0.00 & \textbf{24.47}  & \textbf{34.17} & \textbf{68.34} & 0.00 \\ \svhline 
Train dataset & \multicolumn{8}{c}{IIITD-WVU}   \\ \hline
Test dataset  & \multicolumn{4}{c|}{Notre Dame} & \multicolumn{4}{c}{Clarkson}  \\ \hline
Metric        & EER    & HTER  & APCER & BPCER & EER   & HTER  & APCER & BPCER \\ \hline
DenseNet      & 18.28  & 19.78 & 36.56 & 3.00  & 22.64 & 48.55 & \textbf{0.00}  & 97.10  \\
PBS           & \textbf{12.39}  & \textbf{16.86} & \textbf{33.33} & 0.39  & 37.24 & 47.17 & \textbf{0.00}  & 94.34 \\
APBS          & 15.11  & 27.61 & 54.72 & \textbf{0.33}  & \textbf{21.58} & \textbf{21.95} & 20.80 & \textbf{32.10} \\ \hline
\end{tabular}
\end{center}
\label{tab:cross_db}
\end{table}

\subsubsection{Results on the NDCLD-2013/2015 Database}

Table~\ref{tab:cld_results} compares the iris PAD performance of our models with five state-of-the-art methods on NDCLD-2015 and two different subsets in the NDCLD-2013 database. It can be seen from Table~\ref{tab:cld_results} that our A-PBS model {outperformed all methods} on all databases, revealing the excellent effectiveness of a combination of PBS and attention module on textured contact lens attacks. In addition to comparison with state-of-the-art algorithms, we also report the TDR (\%) at 0.2\% BPCER in Table~\ref{tab:tdr_results_nd_iiit}. {It can be found that despite all three models produce similarly good results, A-PBS obtains slightly better performance than DenseNet and PBS.} The near-perfect results on NDCLD-2013/-2015 databases hint at the obsolescence and limitations of the current iris PAD databases and call for the need for more diversity in iris PAD data.

\begin{table}[htbp!]
\caption{Iris PAD performance of our proposed methods and existing state-of-the-arts on NDCLD-2013/-2015 databases with a threshold of 0.5. The best performance (in terms of lowest HTER) is in bold}
\label{tab:cld_results}
\begin{center}
\resizebox{1.0\textwidth}{!}{%
\begin{tabular}{c|c|c|c|c|c|c||c|c|c}
\hline
\multirow{2}{*}{Database} & \multirow{2}{*}{Metric} & \multicolumn{8}{c}{Presentation Attack Detection Algorithm (\%)} \\ \cline{3-10} 
 & & LBP\cite{lbp14}  & WLBP \cite{wlbp10}  & DESIST \cite{desist16} & MHVF \cite{fusionvgg18} & MSA \cite{DBLP:conf/icb/FangDKK20, DBLP:journals/ivc/FangDBKK21} & DenseNet & PBS & A-PBS \\ \hline
\multirow{3}{*}{NDCLD-2015 \cite{ndcld15}} & ACPER & 6.15 & 50.58 & 29.81 & 1.92 & 0.18 & 1.58 & 1.09 & 0.08 \\ 
& BPCER & 38.70 & 4.41 & 9.22 & 0.39 & 0.00 & 0.14 & 0.00 & 0.06\\
& HTER  & 22.43 & 27.50 & 19.52 & 1.16 & 0.09 & 0.86 & 0.54 & \textbf{0.07} \\ \hline
\multirow{3}{*}{NDCLD13 (LG4000) \cite{ndcld2013}} & APCER & 0.00 & 2.00 & 0.50 & 0.00 & 0.00 & 0.20 & 0.00 & 0.00  \\ 
& BPCER & 0.38 & 1.00 & 0.50 & 0.00 & 0.00 & 0.28 & 0.03 & 0.00  \\
& HTER  & 0.19 & 1.50 & 0.50 & \textbf{0.00} & \textbf{0.00} & 0.24 & 0.02 & \textbf{0.00}\\ \hline 
\multirow{3}{*}{NDCLD13 (AD100) \cite{ndcld2013}} & APCER & 0.00 & 9.00 & 2.00 & 1.00 & 1.00 & 0.00 & 0.00 & 0.00 \\ 
& BPCER & 11.50 & 14.00 & 1.50 & 0.00 & 0.00 & 0.00 & 0.00 & 0.00 \\
& HTER  & 5.75 & 11.50 & 1.75 & 0.50 & 0.50 & \textbf{0.00} & \textbf{0.00} & \textbf{0.00}\\ \hline
\end{tabular}}
\end{center}
\end{table}

\begin{table}[htbp!]
\caption{Iris PAD performance reported in terms of TDR (\%) at 0.2\% BPCER on NDCLD-2013 and NDCLD-2015 databases. The best performance is in bold.}
\label{tab:tdr_results_nd_iiit}
\begin{center}
\begin{tabular}{c|c|c|c}
\hline
\multirow{2}{*}{Database} & \multicolumn{3}{c}{TDR (\%) @ 0.2\% BPCER} \\ \cline{2-4} 
 & DenseNet & PBS & A-PBS \\ \hline
NDCLD-2015 & 99.45 & 99.84 & \textbf{99.96} \\ \hline
NDCLD13 (LG4000) & 99.75 & \textbf{100.00} & \textbf{100.00} \\ \hline
NDCLD13 (AD100) & 100.00 & 100.00 & 100.00 \\ \hline
IIITD-CLI (Cognet) & 99.02 & \textbf{99.59} & 99.57 \\ \hline
IIITD-CLI (Vista) & 100.00 & 100.00 & 100.00 \\ \hline
\end{tabular}
\end{center}
\end{table}

\subsubsection{Results on the IIITD-CLI Database}

\begin{table}[htbp!]
\caption{Iris PAD performance in terms of CCR (\%) on IIITD-CLI database. The best performance is in bold.}
\label{tab:iiit_cld_results}
\begin{center}
\begin{tabular}{c|c|c}
\hline
PAD Algorithms & Cogent & Vista \\
\hline
Textural Features \cite{DBLP:conf/icpr/WeiQST08} & 55.53 & 87.06 \\\hline
WLBP \cite{wlbp10} & 65.40 & 66.91 \\ \hline
LBP+SVM \cite{lbp14} & 77.46 & 76.01 \\ \hline
LBP+PHOG+SVM \cite{DBLP:conf/civr/BoschZM07} & 75.80 & 74.45 \\\hline
mLBP \cite{iiitd_cli} & 80.87 & 93.91 \\ \hline
ResNet18 \cite{DBLP:conf/cvpr/HeZRS16} & 85.15 & 80.97 \\ \hline
VGG \cite{vgg16} & 90.40 & 94.82 \\ \hline
MVANet \cite{Gupta20} & 94.90 & 95.11 \\ \hline \hline
DenseNet & 99.37 & \textbf{100.00} \\\hline
PBS & 99.62 & \textbf{100.00} \\\hline
A-PBS & \textbf{99.70} & \textbf{100.00} \\\hline

\end{tabular}
\end{center}
\end{table}

{Since most of the existing works reported the results using CCR metric on IIITD-CLI database \cite{iiitd_cli, iiitd_cli_2}, we also strictly follow its experimental protocol where we show the experimental results in Table \ref{tab:iiit_cld_results}. In addition to CCR,}  the TDR at 0.2\% BPCER is reported in Table\ref{tab:tdr_results_nd_iiit}. The experiments are performed on Cognet and Vista sensor subsets, respectively. As shown in Table~\ref{tab:cld_results}, our PBS and A-PBS solutions outperform all hand-crafted and CNN-based methods by a large margin (99.79\% on Cognet subset and 100.00\% on Vista subset). The near-perfect classification performance obtained by DenseNet, PBS, and A-PBS reveals that {
despite the significant PAD improvements achieved by deep learning models, there is an urgent need for large-scale iris PAD databases to be built for future research and generalizability analysis.}

{\subsection{Iris PAD in the visible spectrum}}
{In addition to results on NIR databases, we also report results on the visible-light-based PAVID database in Table~\ref{tab:intra-pavid}. The experiments were demonstrated following the defined protocols in \cite{DBLP:conf/btas/RajaRB15a}. For example, the Nokia - iPhone setup refers to the training and testing data as bona fide videos captured using the Nokia phone and the attack videos captured by iPhone. Moreover, we provide the results under a grand test setup, where bona fide and attack data includes videos captured by Nokia and iPhone. The models trained under grand-test setup will be used for cross-spectrum experiments later. It can be observed in Table~\ref{tab:intra-pavid} that deep-learning-based methods, including our A-PBS, outperform all the previously reported results on the PAVID database, which are hand-crafted feature-based PAD solutions. The DenseNet, PBS, and A-PBS methods obtain the best performance with all error rates of 0.00\%.}

\begin{table}[thbp]
\caption{{Iris PAD performance of our proposed methods and established solutions on PAVID database with a threshold of 0.5. The results are reported based on APCER (\%), BPCER (\%), and HTER (\%). Nokia - iPhone refers that the bona fide video is captured by Nokia while the replayed attack video is captured by iphone, and vice versa. Grand Test refers that both, bona fide and reply, videos are captured by Nokia and iphone. The best performance (the lowest HTER value) is in bold.}}
\label{tab:intra-pavid}
\resizebox{1.0\textwidth}{!}{%
\begin{tabular}{c|c|c|c|c|c||c|c|c}
\hline
Video                            & Metric & IQM-SVM \cite{DBLP:journals/tip/GalballyMF14,DBLP:conf/btas/RajaRB15a} & LBP-SVM \cite{DBLP:conf/icb/MaattaHP11,DBLP:conf/btas/RajaRB15a} & BSIF-SVM \cite{DBLP:journals/tifs/RaghavendraB15,DBLP:conf/btas/RajaRB15a} & STFT \cite{DBLP:conf/btas/RajaRB15a} & DenseNet   & PBS        & A-PBS       \\
\hline
\multirow{3}{*}{Nokia - iPhone} & APCER  & 4.50    & 4.51    & 10.81    & 4.46 &0.00         &0.00         &0.00         \\
& BPCER  & 76.92   & 3.84    & 2.56     & 1.28 &0.00         &0.00         &0.00         \\
& HTER   & 40.71   & 4.18    & 6.68     & 2.87 & \textbf{0.00} & \textbf{0.00} & \textbf{0.00} \\
\hline
\multirow{3}{*}{Nokia - Nokia}& APCER  & 3.57    & 2.67    & 0.89     & 2.68 &0.00         &0.00         &0.00         \\
& BPCER  & 57.31   & 4.87    & 6.09     & 1.21 &0.00         &0.00         &0.00         \\
& HTER   & 30.44   & 3.77    & 3.49     & 1.95 & \textbf{0.00} & \textbf{0.00} & \textbf{0.00} \\
\hline
\multirow{3}{*}{iPhone - iPhone} & APCER  & 11.60   & 0.89    & 9.82     & 1.78 &0.00         &0.00         &0.00         \\
& BPCER  & 57.31   & 4.87    & 6.09     & 1.21 &0.00         &0.00         &0.00         \\
& HTER   & 34.45   & 2.88    & 7.96     & 1.49 & \textbf{0.00} & \textbf{0.00} & \textbf{0.00} \\
\hline
\multirow{3}{*}{iPhone - Nokia} & APCER  & 10.71   & 3.54    & 8.92     &0.00   &0.00         &0.00         &0.00         \\
& BPCER  & 76.92   & 3.84    & 2.56     & 1.28 &0.00         &0.00         &0.00         \\
& HTER   & 43.81   & 3.69    & 5.74     & 0.64 & \textbf{0.00} & \textbf{0.00} & \textbf{0.00} \\
\hline
\multirow{3}{*}{Grand-test} & APCER  & -       & -       & -        & -    &0.00         &0.00         &0.00         \\
& BPCER  & -       & -       & -        & -    & 0.00         & 0.00  & 0.00         \\
& HTER   & -       & -       & -        & -    & \textbf{0.00} & \textbf{0.00} & \textbf{0.00} \\
\hline
\end{tabular}}
\end{table}

{\section{Cross-spectrum evaluation results}}

\begin{table}[thbp]
\caption{{Iris PAD performance measured under cross-spectrum scenarios and reported in terms of EER (\%) and HTER (\%), APCER(\%), and BPCER(\%). APCER and BPCER is determined by a threshold of 0.5. The training subset of the grand-test on the visible-light-based PAVID database is used to train a model, and the testing subset of each database in the LivDet-Iris 2017 database is used for evaluation. The lowest error rate is in bold.}}
\label{tab:vis_nir}
\begin{center}
\resizebox{1.0\textwidth}{!}{%
\begin{tabular}{c|cccc|cccc|cccc}
\hline
Train database & \multicolumn{12}{c}{PAVID} \\ \hline
Test database  & \multicolumn{4}{c|}{Clarkson} & \multicolumn{4}{c|}{Notre Dame} & \multicolumn{4}{c}{IIITD-WVU} \\ \hline
Metric & EER & HTER & APCER & BPCER & EER & HTER & APCER & BPCER & EER & HTER & APCER & BPCER \\ \hline
DenseNet & 37.78 & 36.69  & \textbf{45.97} & 27.41 & 56.39 & 56.69 & 59.28 & 54.11 & 54.43 & 49.94 & \textbf{9.40} & 49.94  \\ \hline
PBS      & \textbf{30.43} & 37.12 & 66.23 & \textbf{8.01} & 55.67 & 55.39 & 81.22 & \textbf{29.56} & 51.10 & 50.59 & 82.66 & 18.52 \\ \hline
A-PBS     & 33.41 & \textbf{33.57} & 46.20 & 20.94 & \textbf{53.11} & \textbf{53.83} & \textbf{40.33} & 65.89 & \textbf{26.32} & \textbf{26.13} & 36.30 & \textbf{15.95} \\
\hline
\end{tabular}}
\end{center}
\end{table}

\begin{table}[thbp]
\caption{{Iris PAD performance measured under cross-spectrum scenarios and reported in terms of EER (\%), HTER (\%), APCER(\%), and BPCER(\%). APCER and BPCER is determined by a threshold of 0.5. The training subset of the NIR-based Clarkson, Notre Dame, and IIITD-WVU is used to train a model, and the testing subset of the grand-test on the PAVID database is used for evaluation. The lowest error rate is in bold.}}
\label{tab:nir_vis}
\begin{center}
\resizebox{1.0\textwidth}{!}{%
\begin{tabular}{c|cccc|cccc|cccc}
\hline
Train database & \multicolumn{4}{c|}{Clarkson} & \multicolumn{4}{c|}{Notre Dame} & \multicolumn{4}{c}{IIITD-WVU} \\ \hline
Test database  & \multicolumn{12}{c}{PAVID}  \\ \hline
Metric & EER & HTER & APCER & BPCER & EER & HTER & APCER & BPCER & EER & HTER & APCER & BPCER \\
\hline
DenseNet & 6.04 & 13.53 & 23.94 & 3.13 & 57.49 & 61.40 &95.30 & 27.50 & \textbf{8.28} & 8.07 & \textbf{7.38} & 8.75 \\ \hline
PBS      & 4.47 & \textbf{5.97} & \textbf{10.07} & 1.88 & 56.38 & 57.94 & \textbf{76.51} & 39.38 & 13.43  & 14.15 & 17.67 & 10.63 \\ \hline
A-PBS     & \textbf{1.34} & 12.98 & 25.95 & \textbf{0.00} & \textbf{52.35} & \textbf{50.63} & 100.00 & \textbf{1.25} & 8.63 & \textbf{8.05} & 11.63 & \textbf{5.62} \\
\hline
\end{tabular}}
\end{center}
\end{table}

\begin{figure}[htbp!]
\centering{\includegraphics[width=0.99\linewidth]{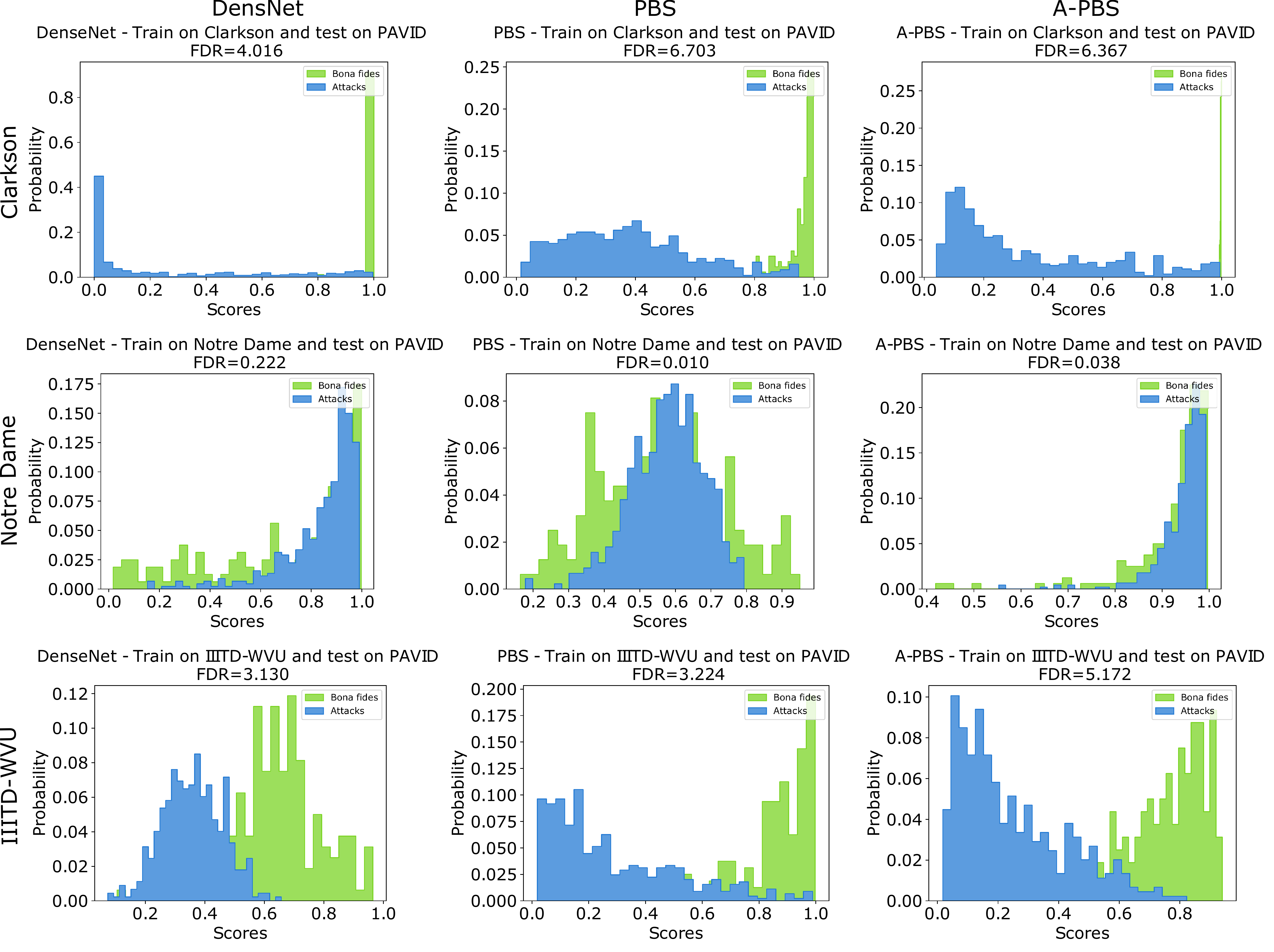}}
\caption{{PAD score distribution of bona fide (green) and PAs (blue) under cross-spectrum scenario (NIR-VIS). The models trained on the training subset of Clarkson (top), Notre Dame (middle), and IIITD-WVU (bottom) databases are used to evaluate on the test subset of PAVID database. and the histograms from left to right are produced by DenseNet, PBS, and A-PBS, respectively. The larger separability (measured by Fisher Discriminant Ratio (FDR) \cite{DBLP:journals/ijon/LorenaC10,DBLP:conf/eusipco/DamerON14}) and smaller overlap indicate higher classification performance.}}
\label{fig:nir_vis_hist}
\end{figure}

\begin{figure}[htbp!]
\centering{\includegraphics[width=0.99\linewidth]{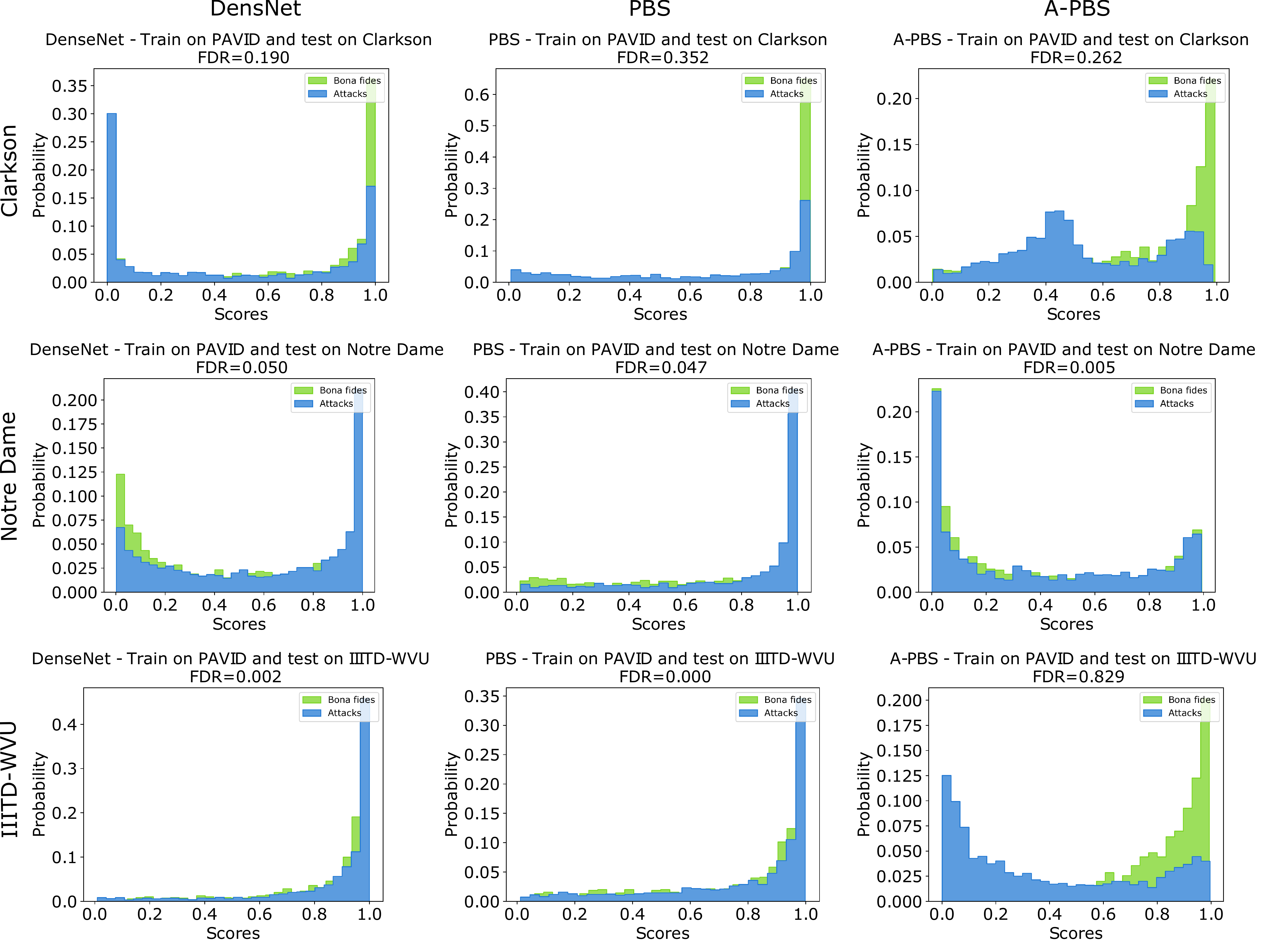}}
\caption{{PAD score distribution of bona fide (green) and PAs (blue) under cross-spectrum scenario (VIS-NIR). The model trained on PAVID database is used to test on the test subset of Clarkson, Notre Dame, and IIITD-WVU, respectively. The histogram top to bottom are test results on Clarkson, Notre Dame and IIITD-WVU databases, and the histograms from left to right are produced by DenseNet, PBS, and A-PBS, respectively. The larger separability (measured by Fisher Discriminant Ratio (FDR) \cite{DBLP:journals/ijon/LorenaC10,DBLP:conf/eusipco/DamerON14}) and smaller overlap indicate higher classification performance.}}
\label{fig:vis_nir_hist}
\end{figure}

{Most studies \cite{fusionvgg18, DBLP:conf/icb/FangDKK20, DBLP:journals/ivc/FangDBKK21, crossdomain19} have presented PAD algorithms and verified their performance on NIR-based database. However, the performance of visible-light iris PAD has been understudied, especially under the cross-spectrum scenario. Therefore, we used the visible-light-based PAVID \cite{DBLP:conf/btas/RajaRB15a} and the NIR-based LivDet-Iris 2017 \cite{livedet17} databases to explore the effect of PAD performance across different spectra. The first scenario is the VIS-NIR where the models trained under the PAVID grand-test setup (visible spectrum) were evaluated on the test subsets of the NIR databses (Clarkson, Notre Dame, and IIIT-WVU), respectively. This evaluation results are presented in Table \ref{tab:vis_nir} and the bold numbers indicate the best performance (lowest error rates). It can be seen that our PBS and A-PBS outperform the trained from scratch DenseNet. However, all PAD methods do not generalize well on the Notre Dame database. One possible reason is that Notre Dame comprises only challenging textured lens attacks and no print/reply attacks.  The PAVID database, used for training here, only include reply attacks. One must note that both reply and print attacks involve the recapture of an artificially presented iris sample, unlike lens attacks. This recapture process can introduce artifacts identifiable by the PAD algorithms. Table \ref{tab:nir_vis} presents the results tested on the PAVID databases by using respective models trained on LivDet-Iris 2017 databases (the case of NIR-VIS). Similar observation can be found in Table \ref{tab:nir_vis} that the model trained on Notre Dame can not generalize on the PAVID database, e.g., the lowest EER and HTER values are 52.34\% and 50.63\% obtained by our A-PBS solution. In contrast to the results on Notre Dame, the model trained on Clarkson and IIITD-WVU generalizes much better on the visible-light database. The lowest EER and HTER values are 1.34\% achieved by A-PBS and 5.97\% achieved by PBS methods, while DenseNet and A-PBS obtained similar error rates on IIITD-WVU. 
Moreover, we illustrate the PAD score distribution with the fisher discriminant ratio \cite{DBLP:journals/ijon/LorenaC10,DBLP:conf/eusipco/DamerON14}, which measures the separability, for further analysis. Figure \ref{fig:nir_vis_hist} and \ref{fig:vis_nir_hist} presents the results of case NIR-VIS and VIR-NIR, respectively. The PAD score distributions of the NIR-VIS case in Figure \ref{fig:nir_vis_hist} show that models trained on Notre Dame perform worse than those trained on Clarkson and IIIT-WVU (bona fide and attack scores almost completely overlap). Moreover, the model trained on PAVID also obtained the largest overlapping and the smallest FDR value in Figure \ref{fig:vis_nir_hist}. One possible reason is the insufficient training data in Notre Dame (1,200 training data). However, the main reason might relate to the type of attacks and the lack of the recapturing process in the lens attacks, as mentioned earlier. This is also verified by the quantitative results in Table \ref{tab:vis_nir} and \ref{tab:nir_vis} (the APCER values are between 40.33\% to 100.00\%).}

\section{Visualization and Explainability}
\begin{figure}[thbp!]
\centering{\includegraphics[width=0.99\linewidth]{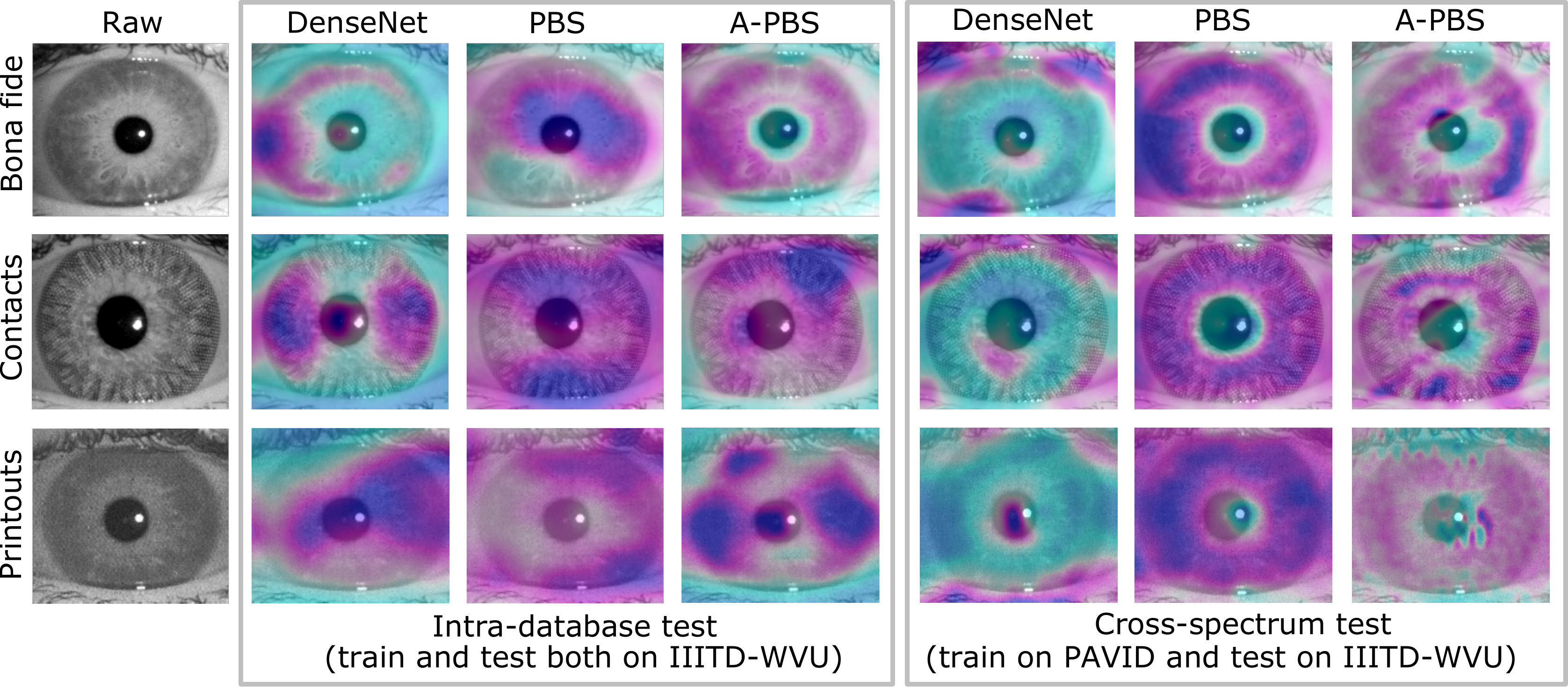}}
\caption{{Score-CAM visualizations for bona fide and attack samples in the IIITD-WVU test set under intra-database and cross-spectrum (VIS-NIR) scenarios. The darker the color of the region, the higher the attention on this area. The column from left to right refers to the raw samples, maps produced by DenseNet, PBS, and A-PBS model under two cases, respectively. The row from top to bottom refers to bona fide samples, textured contact lens, and printouts attack. PBS and A-PBS models pay more attention on iris region than DenseNet in both cases.}}
\label{fig:iiitd_score_cam}
\end{figure}

\begin{figure}[thbp!]
\centering{\includegraphics[width=0.99\linewidth]{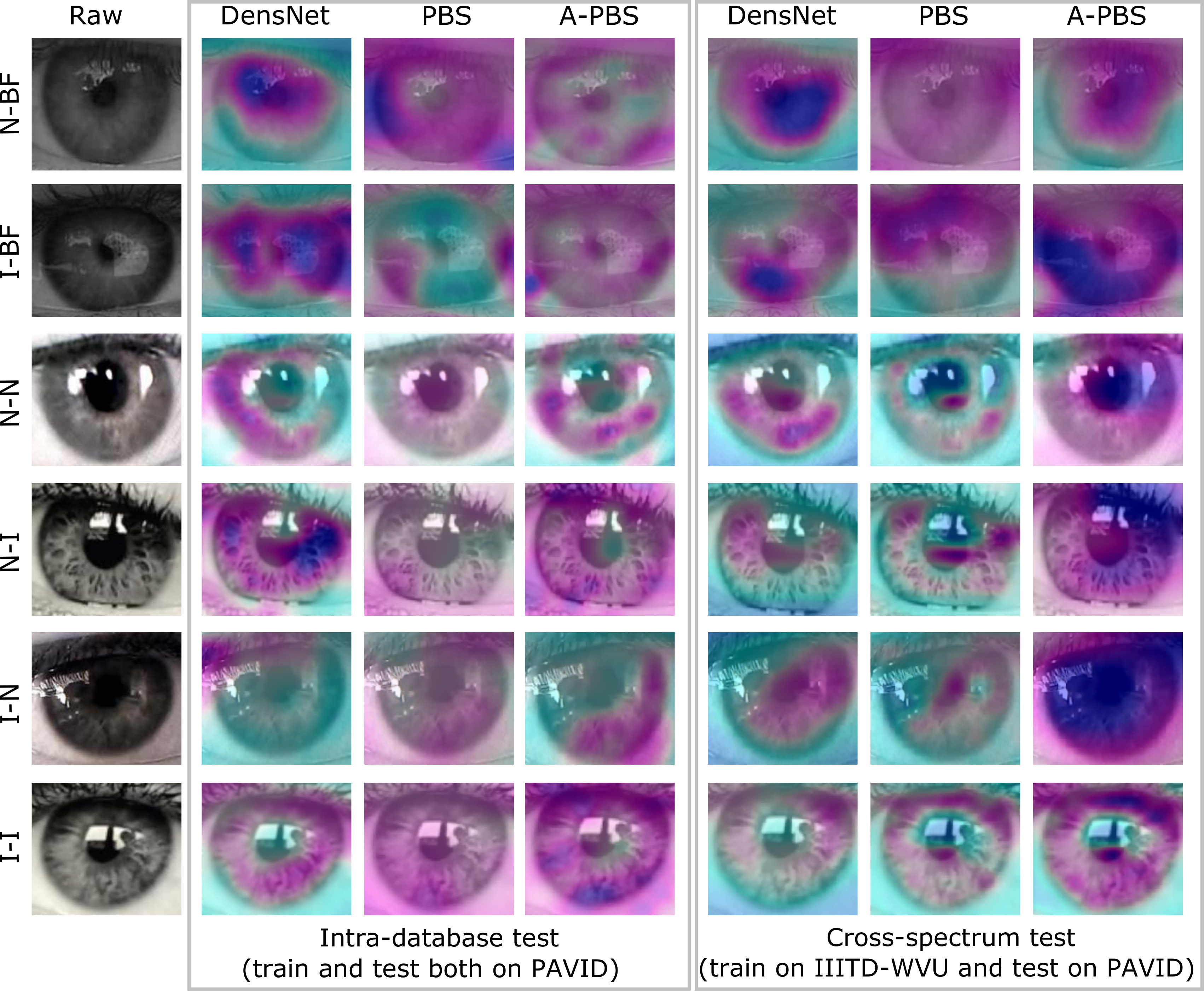}}
\caption{{Score-CAM visualizations for bona fide and attack samples in the PAVID test set under intra-database and cross-spectrum (NIR-VIS) scenarios. The darker the color of the region, the higher the attention on this area. The column from left to right refers to the raw samples, maps produced by DenseNet, PBS, and A-PBS model in two cases, respectively. N and I refer to Nokia and iPhone, respectively. The first two rows refer to the bona fide sample captured by Nokia and iPhone. The following four rows present the replay attack cases: Nokia-Nokia, Nokia-iPhone, iPhone-Nokia, and iPhone-iPhone. It is clear that the A-PBS model with an additional attention module is able to preserve relatively more attention on the iris region, especially for attack samples.}}
\label{fig:pavid_score_cam}
\end{figure}

PBS is expected to learn more discriminative features by supervising each pixel/patch in comparison with binary supervised DenseNet. Subsequently, the A-PBS model, an extended model of PBS, is hypothesized to automatically locate the important regions that carry the features most useful for making an accurate iris PAD decision. To further verify and explain these assumptions, Score-Weighted Class Activation Mapping (Score-CAM) \cite{DBLP:conf/cvpr/WangWDYZDMH20} is used to generate the visualizations for randomly chosen bona fide and attack iris images (these images belong to the same identity) {under intra-database and cross-spectrum scenarios as shown in Figure \ref{fig:iiitd_score_cam} and \ref{fig:pavid_score_cam}}. 

{Figure \ref{fig:iiitd_score_cam} illustrates the score-CAM results on the NIR samples in the test subset of IIITD-WVU. We adopted models trained on the training subset of IIITD-WVU (NIR) and models trained on the training subset of PAVID (visible-light) to generate score-CAMs, respectively. As shown in Figure \ref{fig:iiitd_score_cam}, it is clear that PBS and A-PBS models pay more attention to the iris region than DenseNet in both intra-database and cross-spectrum cases. The DenseNet model seems to lose some information due to binary supervision. Similar observations can be found in Figure \ref{fig:pavid_score_cam}, where the NIR and visible-light models were tested on the visible images in the test subset of the PAVID database. In the visible intra-database case, DenseNet gained more attention on the eye region of visible-light images than of NIR images in Figure \ref{fig:iiitd_score_cam}. Moreover, in the cross-spectrum case in Figure \ref{fig:pavid_score_cam}, the use of the attention module (A-PBS) has enabled the model to keep focusing on the iris area, while DenseNet and PBS lost some attention, especially on the attack samples. In general, the observations in Figures \ref{fig:iiitd_score_cam} and \ref{fig:pavid_score_cam} are consistent with the quantitative results in Table \ref{tab:nir_vis} and \ref{tab:vis_nir} that the training on visible-light and test on NIR data (VIS-NIR) is more challenging than the training on NIR and test on visible-light data (NIR-VIS) in our case. It might be caused by: 1) The perceived image quality of visible data in the PAVID database are relatively lower than NIR images (see samples in Figure \ref{fig:iris_samples}). 2) Some of the video frames in the PAVID database have an eye-blinking process, and thus some iris information (regions) will be hidden by eyelids and eyelashes. 3). While the used visible data (PAVID) contains only recaptured attacks (reply attacks), the NIR data contains both recaptured attacks (print attacks) and lens attacks, which makes it more difficult for a PAD trained on the visible images to perform properly on NIR attacks in our experiments.}

\section{Conclusion} 
\label{sec:con}

{This chapter focuses on the iris PAD performance in the NIR and visible domain, including challenging cross-database and cross-spectrum cases. The experiments were conducted using the novel attention-based pixel-wise binary supervision (A-PBS) method for iris PAD. A-PBS solution aimed to capture the fine-grained pixel/patch-level cues and utilize regions that contribute the most to an accurate PAD decision by utilizing an attention mechanism. The extensive experiments were performed on six publicly available iris PAD databases in the NIR spectrum (including LivDet-Iris 2017 competition databases) and one database in the visible spectrum. By observing intra-database and intra-spectrum experimental results, we concluded that 1) The results reported on respective attack types indicated that textured contact lens attack is more challenging to detect correctly than printouts attack. 2) cross-PA and cross-database are still challenging (EER values are over 20\% in most cases). Furthermore, to our knowledge, this chapter is the first work to perform and analyze experiments under the cross-spectrum scenario. The experimental results showed that models trained on the visible spectrum do not generalize well on NIR data. It might also be caused by the limited visible data and its attack mechanism. In general, the A-PBS solution presents a superior PAD performance and high generalizability in the NIR and visible captured images, cross-database experiments, as well as cross-spectrum PAD deployments. The A-PBS also showed to focus the attention of the PAD models towards the iris region when compared to more traditional solutions. }

\begin{acknowledgement}
This research work has been funded by the German Federal Ministry of Education and Research and the Hessen State Ministry for Higher Education, Research and the Arts within their joint support of the National Research Center for Applied Cybersecurity ATHENE.
\end{acknowledgement}

\ifthenelse{\equal{false}{\buildbook}}{
\printindex
\printglossary
\bibliographystyle{spmpsci}
\bibliography{main}
}

\end{document}

%% file: packages.tex
\usepackage{mathptmx}       
\usepackage{helvet}         
\usepackage{courier}        
\usepackage{type1cm}        
\usepackage{makeidx}         
\usepackage{graphicx}        
\usepackage{multicol}        
\usepackage[bottom]{footmisc}
\usepackage{url}

\usepackage{ifthen}
\usepackage{makeglos}
\usepackage{multirow}
\usepackage{amsmath}
\usepackage{amsfonts}

%% file: main.bbl
\begin{thebibliography}{10}
\providecommand{\url}[1]{{#1}}
\providecommand{\urlprefix}{URL }
\expandafter\ifx\csname urlstyle\endcsname\relax
  \providecommand{\doi}[1]{DOI~\discretionary{}{}{}#1}\else
  \providecommand{\doi}{DOI~\discretionary{}{}{}\begingroup
  \urlstyle{rm}\Url}\fi

\bibitem{DBLP:conf/civr/BoschZM07}
Bosch, A., Zisserman, A., Mu{\~{n}}oz, X.: Representing shape with a spatial
  pyramid kernel.
\newblock In: N.~Sebe, M.~Worring (eds.) Proceedings of the 6th {ACM}
  International Conference on Image and Video Retrieval, {CIVR} 2007,
  Amsterdam, The Netherlands, July 9-11, 2007, pp. 401--408. {ACM} (2007).
\newblock \doi{10.1145/1282280.1282340}.
\newblock \urlprefix\url{https://doi.org/10.1145/1282280.1282340}

\bibitem{DBLP:journals/ivc/Boutros20}
Boutros, F., Damer, N., Raja, K.B., Ramachandra, R., Kirchbuchner, F., Kuijper,
  A.: Iris and periocular biometrics for head mounted displays: Segmentation,
  recognition, and synthetic data generation.
\newblock Image Vis. Comput. \textbf{104}, 104,007 (2020).
\newblock \doi{10.1016/j.imavis.2020.104007}.
\newblock \urlprefix\url{https://doi.org/10.1016/j.imavis.2020.104007}

\bibitem{DBLP:conf/ijcb/Boutros20}
Boutros, F., Damer, N., Raja, K.B., Ramachandra, R., Kirchbuchner, F., Kuijper,
  A.: On benchmarking iris recognition within a head-mounted display for
  {AR/VR} applications.
\newblock In: 2020 {IEEE} International Joint Conference on Biometrics, {IJCB}
  2020, Houston, TX, USA, September 28 - October 1, 2020, pp. 1--10. {IEEE}
  (2020).
\newblock \doi{10.1109/IJCB48548.2020.9304919}.
\newblock \urlprefix\url{https://doi.org/10.1109/IJCB48548.2020.9304919}

\bibitem{DBLP:conf/wacv/ChenR21}
Chen, C., Ross, A.: An explainable attention-guided iris presentation attack
  detector.
\newblock In: {IEEE} Winter Conference on Applications of Computer Vision
  Workshops, {WACV} Workshops 2021, Waikola, HI, USA, January 5-9, 2021, pp.
  97--106. {IEEE} (2021).
\newblock \doi{10.1109/WACVW52041.2021.00015}.
\newblock \urlprefix\url{https://doi.org/10.1109/WACVW52041.2021.00015}

\bibitem{DBLP:conf/eusipco/DamerON14}
Damer, N., Opel, A., Nouak, A.: Biometric source weighting in multi-biometric
  fusion: Towards a generalized and robust solution.
\newblock In: 22nd European Signal Processing Conference, {EUSIPCO} 2014,
  Lisbon, Portugal, September 1-5, 2014, pp. 1382--1386. {IEEE} (2014).
\newblock \urlprefix\url{https://ieeexplore.ieee.org/document/6952496/}

\bibitem{livdet2020}
Das, P., McGrath, J., Fang, Z., Boyd, A., Jang, G., Mohammadi, A., Purnapatra,
  S., Yambay, D., Marcel, S., Trokielewicz, M., Maciejewicz, P., Bowyer, K.W.,
  Czajka, A., Schuckers, S., Tapia, J.E., Gonzalez, S., Fang, M., Damer, N.,
  Boutros, F., Kuijper, A., Sharma, R., Chen, C., Ross, A.: Iris liveness
  detection competition (livdet-iris) - the 2020 edition.
\newblock In: 2020 {IEEE} International Joint Conference on Biometrics, {IJCB}
  2020, Houston, TX, USA, September 28 - October 1, 2020, pp. 1--9. {IEEE}
  (2020).
\newblock \doi{10.1109/IJCB48548.2020.9304941}.
\newblock \urlprefix\url{https://doi.org/10.1109/IJCB48548.2020.9304941}

\bibitem{DBLP:conf/fusion/FangDBKK20}
Fang, M., Damer, N., Boutros, F., Kirchbuchner, F., Kuijper, A.: Deep learning
  multi-layer fusion for an accurate iris presentation attack detection.
\newblock In: {IEEE} 23rd International Conference on Information Fusion,
  {FUSION} 2020, Rustenburg, South Africa, July 6-9, 2020, pp. 1--8. {IEEE}
  (2020).
\newblock \doi{10.23919/FUSION45008.2020.9190424}.
\newblock \urlprefix\url{https://doi.org/10.23919/FUSION45008.2020.9190424}

\bibitem{DBLP:journals/ivc/FangDBKK21}
Fang, M., Damer, N., Boutros, F., Kirchbuchner, F., Kuijper, A.: Cross-database
  and cross-attack iris presentation attack detection using micro stripes
  analyses.
\newblock Image Vis. Comput. \textbf{105}, 104,057 (2021).
\newblock \doi{10.1016/j.imavis.2020.104057}.
\newblock \urlprefix\url{https://doi.org/10.1016/j.imavis.2020.104057}

\bibitem{DBLP:conf/icb/FangDBKK21}
Fang, M., Damer, N., Boutros, F., Kirchbuchner, F., Kuijper, A.: Iris
  presentation attack detection by attention-based and deep pixel-wise binary
  supervision network.
\newblock In: International {IEEE} Joint Conference on Biometrics, {IJCB} 2021,
  Shenzhen, China, August 4-7, 2021, pp. 1--8. {IEEE} (2021).
\newblock \doi{10.1109/IJCB52358.2021.9484343}.
\newblock \urlprefix\url{https://doi.org/10.1109/IJCB52358.2021.9484343}

\bibitem{DBLP:journals/mva/FangDBKK22}
Fang, M., Damer, N., Boutros, F., Kirchbuchner, F., Kuijper, A.: The
  overlapping effect and fusion protocols of data augmentation techniques in
  iris {PAD}.
\newblock Mach. Vis. Appl. \textbf{33}(1), 8 (2022).
\newblock \doi{10.1007/s00138-021-01256-9}.
\newblock \urlprefix\url{https://doi.org/10.1007/s00138-021-01256-9}

\bibitem{DBLP:conf/eusipco/FangDKK20}
Fang, M., Damer, N., Kirchbuchner, F., Kuijper, A.: Demographic bias in
  presentation attack detection of iris recognition systems.
\newblock In: 28th European Signal Processing Conference, {EUSIPCO} 2020,
  Amsterdam, Netherlands, January 18-21, 2021, pp. 835--839. {IEEE} (2020).
\newblock \doi{10.23919/Eusipco47968.2020.9287321}.
\newblock \urlprefix\url{https://doi.org/10.23919/Eusipco47968.2020.9287321}

\bibitem{DBLP:conf/icb/FangDKK20}
Fang, M., Damer, N., Kirchbuchner, F., Kuijper, A.: Micro stripes analyses for
  iris presentation attack detection.
\newblock In: 2020 {IEEE} International Joint Conference on Biometrics, {IJCB}
  2020, Houston, TX, USA, September 28 - October 1, 2020, pp. 1--10. {IEEE}
  (2020).
\newblock \doi{10.1109/IJCB48548.2020.9304886}.
\newblock \urlprefix\url{https://doi.org/10.1109/IJCB48548.2020.9304886}

\bibitem{DBLP:journals/wacv22/Fang22}
Fang, M., Damer, N., Kirchbuchner, F., Kuijper, A.: Learnable multi-level
  frequency decomposition and hierarchical attention mechanism for generalized
  face presentation attack detection.
\newblock In: 2022 {IEEE} Winter Conference on Applications of Computer Vision,
  {WACV} 2022, Hawaii, USA, Jan 04-08, 2022, pp. 3722--3731. {IEEE} Computer
  Society (2022)

\bibitem{DBLP:journals/tip/GalballyMF14}
Galbally, J., Marcel, S., Fi{\'{e}}rrez, J.: Image quality assessment for fake
  biometric detection: Application to iris, fingerprint, and face recognition.
\newblock {IEEE} Trans. Image Process. \textbf{23}(2), 710--724 (2014).
\newblock \doi{10.1109/TIP.2013.2292332}.
\newblock \urlprefix\url{https://doi.org/10.1109/TIP.2013.2292332}

\bibitem{deeppixbis}
George, A., Marcel, S.: Deep pixel-wise binary supervision for face
  presentation attack detection.
\newblock In: 2019 International Conference on Biometrics, {ICB} 2019, Crete,
  Greece, June 4-7, 2019, pp. 1--8. {IEEE} (2019).
\newblock \doi{10.1109/ICB45273.2019.8987370}.
\newblock \urlprefix\url{https://doi.org/10.1109/ICB45273.2019.8987370}

\bibitem{Gupta20}
Gupta, M., Singh, V., Agarwal, A., Vatsa, M., Singh, R.: Generalized iris
  presentation attack detection algorithm under cross-database settings.
\newblock In: 25th International Conference on Pattern Recognition, {ICPR}
  2020, Virtual Event / Milan, Italy, January 10-15, 2021, pp. 5318--5325.
  {IEEE} (2020).
\newblock \doi{10.1109/ICPR48806.2021.9412700}.
\newblock \urlprefix\url{https://doi.org/10.1109/ICPR48806.2021.9412700}

\bibitem{lbp14}
Gupta, P., Behera, S., Vatsa, M., Singh, R.: On iris spoofing using print
  attack.
\newblock In: 22nd International Conference on Pattern Recognition, {ICPR}
  2014, Stockholm, Sweden, August 24-28, 2014, pp. 1681--1686. {IEEE} Computer
  Society (2014).
\newblock \doi{10.1109/ICPR.2014.296}.
\newblock \urlprefix\url{https://doi.org/10.1109/ICPR.2014.296}

\bibitem{DBLP:conf/cvpr/HeZRS16}
He, K., Zhang, X., Ren, S., Sun, J.: Deep residual learning for image
  recognition.
\newblock In: 2016 {IEEE} Conference on Computer Vision and Pattern
  Recognition, {CVPR} 2016, Las Vegas, NV, USA, June 27-30, 2016, pp. 770--778.
  {IEEE} Computer Society (2016).
\newblock \doi{10.1109/CVPR.2016.90}.
\newblock \urlprefix\url{https://doi.org/10.1109/CVPR.2016.90}

\bibitem{densenet}
Huang, G., Liu, Z., van~der Maaten, L., Weinberger, K.Q.: Densely connected
  convolutional networks.
\newblock In: 2017 {IEEE} Conference on Computer Vision and Pattern
  Recognition, {CVPR} 2017, Honolulu, HI, USA, July 21-26, 2017, pp.
  2261--2269. {IEEE} Computer Society (2017).
\newblock \doi{10.1109/CVPR.2017.243}.
\newblock \urlprefix\url{https://doi.org/10.1109/CVPR.2017.243}

\bibitem{ISO301073}
{International Organization for Standardization}: {ISO/IEC DIS 30107-3:2016:
  Information Technology – Biometric presentation attack detection – P. 3:
  Testing and reporting} (2017)

\bibitem{DBLP:journals/prl/JainNR16}
Jain, A.K., Nandakumar, K., Ross, A.: 50 years of biometric research:
  Accomplishments, challenges, and opportunities.
\newblock Pattern Recognit. Lett. \textbf{79}, 80--105 (2016).
\newblock \doi{10.1016/j.patrec.2015.12.013}.
\newblock \urlprefix\url{https://doi.org/10.1016/j.patrec.2015.12.013}

\bibitem{ndcld15}
Jr., J.S.D., Bowyer, K.W.: Robust detection of textured contact lenses in iris
  recognition using {BSIF}.
\newblock {IEEE} Access \textbf{3}, 1672--1683 (2015).
\newblock \doi{10.1109/ACCESS.2015.2477470}.
\newblock \urlprefix\url{https://doi.org/10.1109/ACCESS.2015.2477470}

\bibitem{ndcld2013}
Jr., J.S.D., Bowyer, K.W., Flynn, P.J.: Variation in accuracy of textured
  contact lens detection based on sensor and lens pattern.
\newblock In: {IEEE} Sixth International Conference on Biometrics: Theory,
  Applications and Systems, {BTAS} 2013, Arlington, VA, USA, September 29 -
  October 2, 2013, pp. 1--7. {IEEE} (2013).
\newblock \doi{10.1109/BTAS.2013.6712745}.
\newblock \urlprefix\url{https://doi.org/10.1109/BTAS.2013.6712745}

\bibitem{payeye}
Kentish, P.: Is paying with your iris the future of transactions? polish
  start-up payeye certainly thinks so.
\newblock
  \urlprefix\url{https://emerging-europe.com/business/is\_paying\_with\_your\_iris\_the\_future\_of\_transactions\_polish\_start\_up\_payeye\_certainly\_thinks\_so/}.
\newblock Accessed: 2020-07-07

\bibitem{spoofnet_tuning}
Kimura, G.Y., Lucio, D.R., Jr., A.S.B., Menotti, D.: {CNN} hyperparameter
  tuning applied to iris liveness detection pp. 428--434 (2020).
\newblock \doi{10.5220/0008983904280434}.
\newblock \urlprefix\url{https://doi.org/10.5220/0008983904280434}

\bibitem{iiitd_cli_2}
Kohli, N., Yadav, D., Vatsa, M., Singh, R.: Revisiting iris recognition with
  color cosmetic contact lenses.
\newblock In: J.~Fi{\'{e}}rrez, A.~Kumar, M.~Vatsa, R.N.J. Veldhuis,
  J.~Ortega{-}Garcia (eds.) International Conference on Biometrics, {ICB} 2013,
  4-7 June, 2013, Madrid, Spain, pp. 1--7. {IEEE} (2013).
\newblock \doi{10.1109/ICB.2013.6613021}.
\newblock \urlprefix\url{https://doi.org/10.1109/ICB.2013.6613021}

\bibitem{desist16}
Kohli, N., Yadav, D., Vatsa, M., Singh, R., Noore, A.: Detecting medley of iris
  spoofing attacks using {DESIST}.
\newblock In: 8th {IEEE} International Conference on Biometrics Theory,
  Applications and Systems, {BTAS} 2016, Niagara Falls, NY, USA, September 6-9,
  2016, pp. 1--6. {IEEE} (2016).
\newblock \doi{10.1109/BTAS.2016.7791168}.
\newblock \urlprefix\url{https://doi.org/10.1109/BTAS.2016.7791168}

\bibitem{crossdomain19}
Kuehlkamp, A., da~Silva~Pinto, A., Rocha, A., Bowyer, K.W., Czajka, A.:
  Ensemble of multi-view learning classifiers for cross-domain iris
  presentation attack detection.
\newblock {IEEE} Trans. Inf. Forensics Secur. \textbf{14}(6), 1419--1431
  (2019).
\newblock \doi{10.1109/TIFS.2018.2878542}.
\newblock \urlprefix\url{https://doi.org/10.1109/TIFS.2018.2878542}

\bibitem{DBLP:conf/cvpr/LiuJ018}
Liu, Y., Jourabloo, A., Liu, X.: Learning deep models for face anti-spoofing:
  Binary or auxiliary supervision.
\newblock In: 2018 {IEEE} Conference on Computer Vision and Pattern
  Recognition, {CVPR} 2018, Salt Lake City, UT, USA, June 18-22, 2018, pp.
  389--398. Computer Vision Foundation / {IEEE} Computer Society (2018).
\newblock \doi{10.1109/CVPR.2018.00048}.
\newblock
  \urlprefix\url{http://openaccess.thecvf.com/content\_cvpr\_2018/html/Liu\_Learning\_Deep\_Models\_CVPR\_2018\_paper.html}

\bibitem{DBLP:journals/ijon/LorenaC10}
Lorena, A.C., de~Leon Ferreira~de Carvalho, A.C.P.: Building binary-tree-based
  multiclass classifiers using separability measures.
\newblock Neurocomputing \textbf{73}(16-18), 2837--2845 (2010).
\newblock \doi{10.1016/j.neucom.2010.03.027}

\bibitem{DBLP:conf/icb/MaattaHP11}
M{\"{a}}{\"{a}}tt{\"{a}}, J., Hadid, A., Pietik{\"{a}}inen, M.: Face spoofing
  detection from single images using micro-texture analysis.
\newblock In: 2011 {IEEE} International Joint Conference on Biometrics, {IJCB}
  2011, Washington, DC, USA, October 11-13, 2011, pp. 1--7. {IEEE} Computer
  Society (2011).
\newblock \doi{10.1109/IJCB.2011.6117510}.
\newblock \urlprefix\url{https://doi.org/10.1109/IJCB.2011.6117510}

\bibitem{DBLP:journals/tifs/RaghavendraB15}
Raghavendra, R., Busch, C.: Robust scheme for iris presentation attack
  detection using multiscale binarized statistical image features.
\newblock {IEEE} Trans. Inf. Forensics Secur. \textbf{10}(4), 703--715 (2015).
\newblock \doi{10.1109/TIFS.2015.2400393}.
\newblock \urlprefix\url{https://doi.org/10.1109/TIFS.2015.2400393}

\bibitem{DBLP:conf/btas/RajaRB15}
Raja, K.B., Raghavendra, R., Busch, C.: Iris imaging in visible spectrum using
  white {LED}.
\newblock In: {IEEE} 7th International Conference on Biometrics Theory,
  Applications and Systems, {BTAS} 2015, Arlington, VA, USA, September 8-11,
  2015, pp. 1--8. {IEEE} (2015).
\newblock \doi{10.1109/BTAS.2015.7358769}

\bibitem{DBLP:conf/btas/RajaRB15a}
Raja, K.B., Raghavendra, R., Busch, C.: Presentation attack detection using
  laplacian decomposed frequency response for visible spectrum and
  near-infra-red iris systems.
\newblock In: {IEEE} 7th International Conference on Biometrics Theory,
  Applications and Systems, {BTAS} 2015, Arlington, VA, USA, September 8-11,
  2015, pp. 1--8. {IEEE} (2015).
\newblock \doi{10.1109/BTAS.2015.7358790}.
\newblock \urlprefix\url{https://doi.org/10.1109/BTAS.2015.7358790}

\bibitem{samsung_iris_scanner}
SAMSUNG ELECTRONICS~CO., L.: How does the iris scanner work on galaxy s9,
  galaxy s9+, and galaxy note9?
\newblock
  \urlprefix\url{https://www.samsung.com/global/galaxy/what-is/iris-scanning/}.
\newblock Accessed: 2021-04-19

\bibitem{DBLP:conf/icb/SharmaR20}
Sharma, R., Ross, A.: D-netpad: An explainable and interpretable iris
  presentation attack detector.
\newblock In: 2020 {IEEE} International Joint Conference on Biometrics, {IJCB}
  2020, Houston, TX, USA, September 28 - October 1, 2020, pp. 1--10. {IEEE}
  (2020).
\newblock \doi{10.1109/IJCB48548.2020.9304880}.
\newblock \urlprefix\url{https://doi.org/10.1109/IJCB48548.2020.9304880}

\bibitem{vgg16}
Simonyan, K., Zisserman, A.: Very deep convolutional networks for large-scale
  image recognition.
\newblock In: Y.~Bengio, Y.~LeCun (eds.) 3rd International Conference on
  Learning Representations, {ICLR} 2015, San Diego, CA, USA, May 7-9, 2015,
  Conference Track Proceedings (2015).
\newblock \urlprefix\url{http://arxiv.org/abs/1409.1556}

\bibitem{DBLP:conf/cvpr/WangWDYZDMH20}
Wang, H., Wang, Z., Du, M., Yang, F., Zhang, Z., Ding, S., Mardziel, P., Hu,
  X.: Score-cam: Score-weighted visual explanations for convolutional neural
  networks.
\newblock In: 2020 {IEEE/CVF} Conference on Computer Vision and Pattern
  Recognition, {CVPR} Workshops 2020, Seattle, WA, USA, June 14-19, 2020, pp.
  111--119. Computer Vision Foundation / {IEEE} (2020).
\newblock \doi{10.1109/CVPRW50498.2020.00020}.
\newblock
  \urlprefix\url{https://openaccess.thecvf.com/content\_CVPRW\_2020/html/w1/Wang\_Score-CAM\_Score-Weighted\_Visual\_Explanations\_for\_Convolutional\_Neural\_Networks\_CVPRW\_2020\_paper.html}

\bibitem{DBLP:conf/icpr/WeiQST08}
Wei, Z., Qiu, X., Sun, Z., Tan, T.: Counterfeit iris detection based on texture
  analysis.
\newblock In: 19th International Conference on Pattern Recognition {(ICPR}
  2008), December 8-11, 2008, Tampa, Florida, {USA}, pp. 1--4. {IEEE} Computer
  Society (2008).
\newblock \doi{10.1109/ICPR.2008.4761673}.
\newblock \urlprefix\url{https://doi.org/10.1109/ICPR.2008.4761673}

\bibitem{cbam}
Woo, S., Park, J., Lee, J., Kweon, I.S.: {CBAM:} convolutional block attention
  module.
\newblock In: V.~Ferrari, M.~Hebert, C.~Sminchisescu, Y.~Weiss (eds.) Computer
  Vision - {ECCV} 2018 - 15th European Conference, Munich, Germany, September
  8-14, 2018, Proceedings, Part {VII}, \emph{Lecture Notes in Computer
  Science}, vol. 11211, pp. 3--19. Springer (2018).
\newblock \doi{10.1007/978-3-030-01234-2\_1}.
\newblock \urlprefix\url{https://doi.org/10.1007/978-3-030-01234-2\_1}

\bibitem{fusionvgg18}
Yadav, D., Kohli, N., Agarwal, A., Vatsa, M., Singh, R., Noore, A.: Fusion of
  handcrafted and deep learning features for large-scale multiple iris
  presentation attack detection.
\newblock In: 2018 {IEEE} Conference on Computer Vision and Pattern Recognition
  Workshops, {CVPR} Workshops 2018, Salt Lake City, UT, USA, June 18-22, 2018,
  pp. 572--579. Computer Vision Foundation / {IEEE} Computer Society (2018).
\newblock \doi{10.1109/CVPRW.2018.00099}.
\newblock
  \urlprefix\url{http://openaccess.thecvf.com/content\_cvpr\_2018\_workshops/w11/html/Yadav\_Fusion\_of\_Handcrafted\_CVPR\_2018\_paper.html}

\bibitem{iiitd_cli}
Yadav, D., Kohli, N., Jr., J.S.D., Singh, R., Vatsa, M., Bowyer, K.W.:
  Unraveling the effect of textured contact lenses on iris recognition.
\newblock {IEEE} Trans. Inf. Forensics Secur. \textbf{9}(5), 851--862 (2014).
\newblock \doi{10.1109/TIFS.2014.2313025}.
\newblock \urlprefix\url{https://doi.org/10.1109/TIFS.2014.2313025}

\bibitem{DBLP:conf/icb/YadavKVSN17}
Yadav, D., Kohli, N., Vatsa, M., Singh, R., Noore, A.: Unconstrained visible
  spectrum iris with textured contact lens variations: Database and
  benchmarking.
\newblock In: 2017 {IEEE} International Joint Conference on Biometrics, {IJCB}
  2017, Denver, CO, USA, October 1-4, 2017, pp. 574--580. {IEEE} (2017).
\newblock \doi{10.1109/BTAS.2017.8272744}.
\newblock \urlprefix\url{https://doi.org/10.1109/BTAS.2017.8272744}

\bibitem{densepad19}
Yadav, D., Kohli, N., Vatsa, M., Singh, R., Noore, A.: Detecting textured
  contact lens in uncontrolled environment using densepad.
\newblock In: {IEEE} Conference on Computer Vision and Pattern Recognition
  Workshops, {CVPR} Workshops 2019, Long Beach, CA, USA, June 16-20, 2019, pp.
  2336--2344. Computer Vision Foundation / {IEEE} (2019).
\newblock \doi{10.1109/CVPRW.2019.00287}.
\newblock
  \urlprefix\url{http://openaccess.thecvf.com/content\_CVPRW\_2019/html/Biometrics/Yadav\_Detecting\_Textured\_Contact\_Lens\_in\_Uncontrolled\_Environment\_Using\_DensePAD\_CVPRW\_2019\_paper.html}

\bibitem{livedet17}
Yambay, D., Becker, B., Kohli, N., Yadav, D., Czajka, A., Bowyer, K.W.,
  Schuckers, S., Singh, R., Vatsa, M., Noore, A., Gragnaniello, D., Sansone,
  C., Verdoliva, L., He, L., Ru, Y., Li, H., Liu, N., Sun, Z., Tan, T.: Livdet
  iris 2017 - iris liveness detection competition 2017.
\newblock In: 2017 {IEEE} International Joint Conference on Biometrics, {IJCB}
  2017, Denver, CO, USA, October 1-4, 2017, pp. 733--741. {IEEE} (2017).
\newblock \doi{10.1109/BTAS.2017.8272763}.
\newblock \urlprefix\url{https://doi.org/10.1109/BTAS.2017.8272763}

\bibitem{wlbp10}
Zhang, H., Sun, Z., Tan, T.: Contact lens detection based on weighted {LBP}.
\newblock In: 20th International Conference on Pattern Recognition, {ICPR}
  2010, Istanbul, Turkey, 23-26 August 2010, pp. 4279--4282. {IEEE} Computer
  Society (2010).
\newblock \doi{10.1109/ICPR.2010.1040}.
\newblock \urlprefix\url{https://doi.org/10.1109/ICPR.2010.1040}

\end{thebibliography}
